\documentclass{CVM}
\newcommand{\eg}{\emph{e.g.}}
\newcommand{\ie}{\emph{i.e.}}
\newcommand{\fzz}[1]{{\color{black}{#1}}}
\newcommand{\whj}[1]{\color{black}{#1}}
\newcommand{\trafps}{\emph{TrafPS}}
\newcommand{\sect}{Sect.~}
\newcommand{\regionshap}{\emph{region SHAP}}
\newcommand{\trajectoryshap}{\emph{trajectory SHAP}}
\newcommand{\fig}{Fig.~}
\newcommand{\eq}{Eq.~}

\newtheorem{theorem}{Theorem}[section]
\CVMsetup{
type      = {Research/Review Article},
doi       = {s41095-0xx-xxxx-x},
title     = {\trafps: A Shapley-based Visual Analytics Approach to Interpret Traffic},
author    = {Zezheng Feng$^{1,2}$, Yifan Jiang$^{1}$, Hongjun Wang$^{1}$, Zipei Fan$^{4}$\cor{}, Yuxin Ma$^{1}$, Shuang-Hua Yang$^{5,3,1}$, Huamin Qu$^{2}$, and Xuan Song$^{1}$\cor{}\\
},
runauthor = {Z. Feng, Y. Jiang, H. Wang, Z. Fan, Y. Ma, SH. Yang, H. Qu, X. Song},
abstract  = {
   Recent achievements in deep learning (DL) have shown its potential for predicting traffic flows. Such predictions are beneficial for understanding the situation and making decisions in traffic control. However, most state-of-the-art DL models are considered ``black boxes'' with little to no transparency for end users with respect to the underlying mechanisms. Some previous work tried to ``open the black boxes'' and increase the interpretability of how predictions are generated. However, it still remains challenging to handle complex models on large-scale spatio-temporal data and discover salient spatial and temporal patterns that significantly influence traffic flows. To overcome the challenges, we present \trafps, a visual analytics approach for interpreting traffic prediction outcomes to support decision-making in traffic management and urban planning. The measurements, \regionshap~and \trajectoryshap, are proposed to quantify the impact of flow patterns on urban traffic at different levels. Based on the task requirement from the domain experts, we employ an interactive visual interface for multi-aspect exploration and analysis of significant flow patterns. Two real-world case studies demonstrate the effectiveness of \trafps~in identifying key routes and decision-making support for urban planning. 

},
keywords  = {Data Visualization, Model Interpretation, Urban Planning, Urban Visual Analytics},
copyright = {The Author(s)},
}

\begin{document}

\maketitle

    \begin{figure}[b] \vskip -4mm
    \small\renewcommand\arraystretch{1.3}
        \begin{tabular}{p{80.5mm}} \toprule\\ \end{tabular}
        \vskip -4.5mm \noindent \setlength{\tabcolsep}{1pt}
        \begin{tabular}{p{3.5mm}p{80mm}}
    $1\quad $ & Department of Computer Science and Engineering, Southern University of Science and Technology. \\
    $2\quad $ & Department of Computer Science and Engineering, Hong Kong University of Science and Technology. \\
    $3\quad $ & Department of Computer Science, University of Reading. \\
    $4\quad $ & Center for Spatial Information Science, The University of Tokyo.\\
    $5\quad $ & Shenzhen Key Laboratory of Safety and Security for Next Generation of Industrial Internet, Southern University of Science and Technology\\
    \cor{} & ~The corresponding author: Xuan Song. songx@sustech.edu.cn
    Zipei Fan. fanzipei@iis.u-tokyo.ac.jp\\
  
&\hspace{-5mm} Manuscript received: 2022-01-01; accepted: 2022-01-01\vspace{-2mm}
    \end{tabular} \vspace {-3mm}
    \end{figure}
% % % % % % % % % % % % % % % % % % % % % % % % % % % % % % % % % % % % % % % % % % % % % % % % % % % % % % % % % % 

\section{Introduction}

Traffic flow prediction is an essential task in traffic data analysis. Real-time and accurate predictions can benefit stakeholders and decision-makers when responding to potential traffic congestion or accidents~\cite{zeng2020revisiting,pi2019visual}. To date, much of the previous work has been proposed to increase the accuracy, time efficiency, and usability of prediction outcomes. Earlier approaches focused on building regression models to fit the traffic flows, while deep learning-based techniques, as the most significant representatives in recent years, have been brought under the limelight in the recent decade due to their strong forecasting performance on traffic flows~\cite{nagy2018survey}.

However, the previous research on deep learning techniques exhibits limitations of lacking transparency. By leveraging state-of-art deep learning models, the studies~\eg,~\cite{pi2019visual,lee2019visual} only focus on the prediction task regardless of the ``black box'' issue with little to no interpretation to the domain experts. \fzz{Without the transparency of the prediction model, the analysts cannot understand the cause of traffic congestion in the near future, and they have no idea about making proper decisions on traffic management and planning the urban infrastructures to improve urban traffic. Furthermore, as the urban planning and traffic management tasks are usually complex from supporting multi-criteria decisions~\cite{liu2016smartadp} due to the uncertainties of the urban environment~\cite{feng2022survey}, the roles of the end user (analysts in urban traffic management and urban planning) are to supply sufficient evidence to the decision makers (the administrator of the city who may not have enough domain knowledge) for making decisions, so the end user should provide an intuitive, easy understand the evidence and make clear interpretation to the “black box” models. Thus, the existing research may be limited, \eg, the analytical tools~\cite{guo2011tripvista,piringer2012alvis} cannot handle the scale of a large urban area for supporting traffic management and analytical tasks.}

% challenges
For accurately and intuitively interpreting the traffic flow prediction in the urban area, we propose this research which focuses on providing urban analysts with an intuitive interpretation of urban traffic, specifically in traffic congestion prediction and urban traffic planning. We distill the consideration of this work as follows:

\vspace{0.4em}
\fzz{\noindent\textbf{Quantifying the Impact on the Traffic from Surroundings.}~The traffic flow in a region is the summary of the trajectory passing through it in a certain period, which may come from different regions around the selected region. Moreover, for a detailed analysis of the urban traffic, which road/routes affect the urban traffic more is a key factor for making decisions. Therefore, it is urgent to propose a method for quantifying the impact on the traffic from surroundings, that is, to measure the contribution to urban traffic from region to region and road/routes to the region, respectively.}

% \vspace{0.4em}
% \noindent\textbf{Revealing the contribution to the traffic from surroundings.}~The traffic flow in a region is the summary of the trajectory passing through it in a certain period, which may come from different regions around the selected region. Moreover, for a detailed analysis of the urban traffic, which road/routes affect the urban traffic more is a key factor for making decisions. Therefore, it is urgent to propose a method for measuring the contribution to urban traffic from region to region and road/routes to the region, respectively.

\vspace{0.4em}
\fzz{\noindent\textbf{Interpreting the Traffic from Spatial and Temporal Dimensions.}~Either quantifying the regional or the road/route contribution to the urban traffic is based on the spatial dimension, while the prediction from deep learning models is a temporal result. Analyzing only one dimension may have a bias that cannot comprehensively understand traffic prediction. Moreover, combining this two-dimensional information efficiently and intuitively is also challenging. So, a visualization technique is needed for a comprehensive analysis from different dimensions.}

\vspace{0.4em}
\fzz{\noindent\textbf{Supporting Multi-level Analysis.}~Demonstrating all levels of visualizations together is a complicated task. Instead, a step-by-step analysis enables the analysts who do not have enough domain knowledge on deep learning to gain different information and understand the model result at each level, \eg, region to region and road/route to the region.}

% Model Framework
The visual analytics approach has been proven an effective tool for analyzing urban traffic~\cite{chen2015survey}. To solve this problem, inspired by previous urban visual analytics studies (\eg,~\cite{pi2019visual,weng2020towards,he2019interactive}), we propose a visual analytics approach named \trafps, aiming to support the interpretation of the urban traffic prediction. The \trafps~contains three layers with different functions to provide efficient analysis: i) Data processing layer preprocesses raw data, partitions the urban area into grids for prediction, and aggregates these grids for interpretation. ii) Prediction-interpretation layer consists of two computation models. The prediction model predicts the traffic flow, and the interpretation model provides the corresponding interpretation to the user. iii) Visualization analytical layer visualizes results in former layers through several interactive views to provide the visualization-assisted analysis (\eg, Map-trajectory View, Radar Glyph View, and Fine-grained Grid View).

% contributions
The contributions of this work are as follows:
\begin{itemize}
\item Two measurements named the \regionshap~and the \trajectoryshap, respectively, to support revealing the impact of the urban traffic at different levels, are proposed by introducing the Shapley value to the traffic interpretation task.

\item An visual analytics system named \trafps~with newly designed visualizations to support multi-level analysis in understanding the prediction and help users make effective decisions. 

\item Two real-world case studies, and interviews from the domain experts to demonstrate the feasibility, usability, and effectiveness of our proposed approach. 

\end{itemize}

% % % % % % % % % % % % % % % % % % % % % % % % % % % % % % % % % % % % % % % % % % % % % % % % % % % % % % % % % % % % % % % % % % 

\section{Related Work}\label{sec: related work}

We summarize the related studies in three parts, including visual analytics for traffic data~(\sect\ref{sec: va for traffic data}), visual interpretation for urban traffic~(\sect\ref{sec: va for traffic}), and explainable AI approach for traffic prediction~(\sect\ref{sec: xai for traffic}).
%------------------------------------------------------------------------------
\subsection{Visual Analytics for Traffic Data}\label{sec: va for traffic data}
According to the fruitful advanced location-sensing technologies for collocating ubiquitous and abundant movement data~\cite{short2010nonlinear}, which records the status of moving objects from both spatial and temporal aspects~\cite{kraak2020cartography}, many researchers have focused on analyzing the insights from these data, especially the urban traffic data.
% [ZEZHENG: THIS SENTENCE IS MALFORMED. PLEASE REWRITE THIS. ALSO, DON'T USE "THANKS TO".] 
Urban traffic data records the status of moving objects (e.g., vehicles and people) in an urban area and has the potential value of reflecting the moving behaviors of the objects~\cite{feng2022survey}. Many researchers have used various kinds of traffic data in different cities to do visual analytics for different purposes, such as supporting exploration for forecasting and monitoring the traffic congestion through loop sensor data in Ulsan~\cite{lee2019visual}, improving the performance of bus route planning through bus trip data in Beijing~\cite{weng2020towards}, providing the suitable solution for billboard location through taxi trajectory data in Tianjin~\cite{liu2016smartadp}, capturing and interpreting the propagation process of the air pollution through the urban environment data~\cite{deng2019airvis}, and analyzing the mobility patterns for the human through taxi trajectory data in New York City~\cite{ferreira2013visual}. Other works can be referred to in the surveys~\cite{andrienko2017visual,chen2015survey,zheng2016visual, feng2022survey}.

Urban traffic visual analytics can be categorized into four stages: descriptive analytics in urban visualization, diagnostic analytics in urban visualization, predictive analytics in urban visualization, and prescriptive analytics in urban visualization~\cite{feng2022survey}. Descriptive analytics in the urban area aims to improve the visualization techniques for visualizing massive urban data~\cite{kamw2019urban,feng2020topology,zeng2019route,zeng2014visualizing} and optimize the traditional urban data mining method through visualization~\cite{kruger2018visual,al2016semantictraj,huang2019natural}. Diagnostic visual analytics supports the tasks for urban pattern exploration~\cite{wu2015telcovis,shen2017streetvizor,qu2007visual} and urban situation surveillance~\cite{cao2017voila,pi2019visual} through visualization. The representative task in predictive visual analytics in the urban area is explainable AI (XAI)~\cite{lee2019visual} aiming to provide an intuitive, effective, and interactive visual analytics approach for prediction model interpretation. And prescriptive analytics in urban visualization can support for location selection~\cite{weng2018homefinder,li2020warehouse,liu2016smartadp} and route planning~\cite{di2015allaboard,weng2020pareto,weng2020towards} which aims to provide comprehensive support for decision-making.

\trafps~can be seen as a predictive visual analytics approach for analyzing the trajectory data that provide interpretation to the data-driven traffic prediction models.

\subsection{Visual Interpretation for Urban Traffic}\label{sec: va for traffic}

In recent years, combining visual analytics approach to interpret the ``black box'' models has been a trend~\cite{ming2017understanding} which has been used to provide understanding~\cite{ma2019explaining,ma2020visual}, diagnosis~\cite{zeng2020revisiting}, and reasoning~\cite{chen2017vaud} for the complex models~\cite{abdul2018trends,liu2017towards}. The traditional approaches for traffic prediction tackle individual variables, \eg, the average speed of the road, the density of a certain area in one period, or counting the number of vehicles. Such approaches are easy to understand by a single number or the basic visual techniques (\eg, bar chart, line chart), and there is no need to interpret further. However, as the ubiquitous urban data are collected and the fruitful demand of the analysis on the urban traffic, the researchers found that the deep learning models act as an important role in good performance so that they are improved to predict the traffic, \eg, DGCN~\cite{guo2020dynamic}, ST-ResNet~\cite{zhang2017deep}. While with the complexity of the deep learning model and the increase in real-world demand, it is necessary to increase the transparency of the prediction, which lets the user trust the ``black-box" deep learning model. Visual analytics approaches have been studied as an effective approach for increasing transparency and interpreting the black-box model on traffic prediction. 

In urban areas, interpreting and understanding traffic status is the precondition, especially for the prediction of short-term traffic prediction~\cite{pi2019visual}. As mentioned above, basic visualization techniques can be used for interpreting single variables. However, these variables lack diversity which may cause high uncertainty in the interpretation. Therefore, for the complicated traffic prediction problem, a comprehensive visual analytics approach or multi-view dashboard has been used in recent research. Wang et al.,~\cite{wang2013visual} proposed an interactive visual analytics tool and newly developed traffic jam graph to analyze the traffic jam and help explain the propagation. to investigate the relations between the urban traffic data and deep learning model through unit visualization techniques, Zeng et al.,~\cite{zeng2020revisiting} analyzed the modifiable areal unit problem (MAUP) that may cause the perturbations for the prediction and proposed a visual analytic solution to depict the input traffic data and the errors for the prediction at different scales. Lee et al.,~\cite{lee2019visual} developed an interactive visual analytics approach with newly designed visualizations to predict and explore the future traffic status on the road. To further explore the cause of traffic congestion, Pi et al.,~\cite{pi2019visual} proposed a novel visual analytics approach that supports the analysts to explore the cause and understand the future predicted traffic congestion. For interpreting the urban traffic and then supporting decision making, RCMVis.,~\cite{shin2021rcmvis} studies the route choice problem by explaining the route choice model first and then helping refine the model. In addition, VATLD~\cite{gou2020vatld} applies disentangled representation learning and semantic adversarial learning to help interpret and understand traffic light detection. Moreover, VASS~\cite{he2021can} utilizes the visual analytics approach to support and diagnose the robustness of semantic segmentation models, which improves the driving sense in urban traffic.

% [ZEZHENG: WE NEED TO IDENTIFY WHERE TRAFPS LOCATES HERE.]

In \trafps, we design a novel visualization technique named the radar glyph chart, which is covered on the map to integrate the prediction and interpretation results together.

\begin{figure*}[htbp]
\includegraphics[width=1\linewidth]{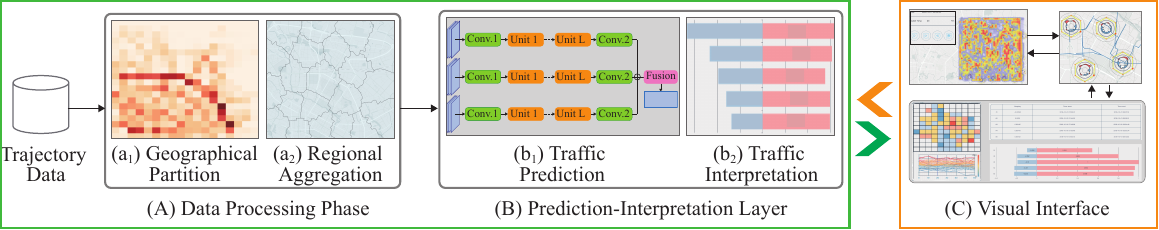}

   \caption{The overview of \trafps. \trafps~consists of three parts, including the (A) data processing phase, (B) prediction interpretation layer, and (C) visual interface. The input of \trafps~is vehicle trajectory data and road network data.}
\label{fig: overview}
\end{figure*}

\subsection{Explainable AI Approach for Traffic Prediction}\label{sec: xai for traffic}

Explainable AI (XAI) can assist the understanding of learning and prediction processes in deep learning models~\cite{doshi2017towards}. Arrieta et al.,~\cite{arrieta2020explainable} divided the methodologies of XAI into two categories: explanation of transparent machine learning models, and post-hoc explanation techniques for machine learning models. In this work, we mainly focus on the techniques for the post-hoc explanation. Several advanced methods have been developed to give the predictions of the model an accurate and comprehensive explanation. These methods can be categorized into model-related methods and model-agnostic methods. Model-agnostic methods explain predictions by checking the relation between input and prediction results, which is agnostic to the task and the model details~\cite{ribeiro2018anchors}. Several model-agnostic methods have been widely used and achieved good effects, \eg, LIME and Shapley value. Riberio et al.,~\cite{ribeiro2016should} introduced LIME which provides the explanation of any predictions used in a faithful way based on approximating it locally with an interpretable model.

The Shapley value applied in gaming theory is a suitable method for providing a model-agnostic explanation of the model prediction~\cite{lundberg2017unified}. It is considered a unique unbiased approach to fairly allocating the total award of a coalition to each input feature~\cite{zhang2021building}. Unlike recent model-agnostic explanation methods~\cite{ribeiro2016should,sundararajan2017axiomatic}, estimation methods based on the Shapley value are better aligned with human intuition and effectually discriminate among model output classes. In the area of traffic prediction, the Shapley value is applied to explain models to give a better comprehension of the predictions, such as to interpret the eXtreme Gradient Boosting (XGBoost) when detecting the potential traffic accidents~\cite{parsa2020toward}, to find the impact on the changes of the features in average daily traffic (ADT)~\cite{parsa2021data}, and to estimate the effect between the agent selected phase and the road detectors state when explaining the reinforcement learning method for traffic signal control~\cite{rizzo2019reinforcement}.

We develop \trafps~that applies the Shapley value from the spatial and temporal dimension to provide intuitive interpretation at a multi-scale (from region to region interpretation to region to trajectory interpretation). In addition, two optimized Shapley value representation methods namely \regionshap~and \trajectoryshap~are proposed to facilitate the interpretation of predicted traffic flow.

% % % % % % % % % % % % % % % % % % % % % % % % % % % % % % % % % % % % % % % % % % % % % % % % % % % % % %

\section{Task Analysis and Solution Overview}\label{sec: requirement and overview}

\subsection{Task Analysis}\label{sec: task}
This research was conducted in close collaboration with four domain experts in traffic data analysis. One is the expert \textbf{E.B} in~\sect\ref{sec: expert interview}, while the other three are experienced experts in the urban management field.
All the collaborators have at least 7 years of experience in urban traffic planning and emergency response management. We held weekly meetings with the collaborators in the early stage of this research. During the meetings, the collaborators helped us understand the data and invited us to attend their regular meetings. We referred to cause and effect analysis~\cite{ishikawa1990introduction} for the problems
% we faced~\cite{sun2020dfseer}.
% [ZEZHENG: STILL NOT MAKE SENSE TO ME. DELETE IT?]. 
and then, summarized two main tasks based on collaborators' requirements and the exploration of related literature.

\vspace{0.4em}
\noindent\fzz{\textbf{T.1 Understanding the Traffic Congestion and Identifying Key Routes.}}~The first task focuses on identifying key routes which may influence the traffic. During the interpretation of the traffic, despite identifying when the traffic congestion exists, it is also a key problem on which area or road may positively impact the selected region or road at a different future period. Therefore, the analysts want to know: \textit{How is the traffic condition in the urban area?} \textit{What will the traffic condition be in the following $t$ period?} \textit{Which surrounding region or roads contribute to the congestion for the selected region?}

\vspace{0.4em}
\noindent\fzz{\textbf{T.2 Understanding the Cause of the Congestion and Supporting Decision-making.}~The second task is to help urban planners in their decision-making process by understanding the cause of traffic congestion. For example, when planning traffic infrastructures to improve urban traffic \ie, how to control reversible lanes.} So they may need to understand: \textit{Which directions have a positive impact on traffic congestion and which have a negative impact?} \textit{Which routes affect more to the urban traffic in the selected region?} \textit{How long will the selected route affects the traffic?}

% \vspace{0.4em}
% \noindent\textbf{T.1 Identify Key Routes in the Traffic Congestion.}~The first task focuses on identifying key routes which may influence the traffic. During the interpretation of the traffic, despite identifying when the traffic congestion exists, it is also a key problem on which area or road may positively impact the select region or road at a different future period. Therefore, the analysts want to know: \textit{How is the traffic condition in the urban area?} \textit{What will the traffic condition be in the following $t$ period?} \textit{Which surrounding region or roads contribute to the congestion for the selected region?}

% \vspace{0.4em}
% \noindent\textbf{T.2 Support Decisions for improving the urban traffic.}~The second task is to help urban planners in their decision-making process, which supports improving urban traffic. For example, when planning traffic infrastructures to improve urban traffic, urban planners usually need to consider how to control reversible lanes. So they may need to understand: \textit{Which directions have a positive impact on traffic congestion and which have a negative impact?} \textit{Which routes affect more to the urban traffic in the selected region?} \textit{How long will the selected route affect the traffic?}

\subsection{Solution Overview}
Based on the key features summarized in the task analysis stage, we design a visual analytics approach, \trafps, to interpret and explore the predictions from deep learning models. The approach consists of three modules: data processing, prediction-interpretation layer, and visual analytical interface, \fig\ref{fig: overview} illustrates the overview of \trafps. 
% [ZEZHENG: MOVE THE TWO DATA SOURCE COMMENTS INTO FOOTNOTES].

% [ZEZHENG: PLEASE RE-ORGANIZE THE COMMENTED TWO PARAGRAPHS INTO THE FOLLOWING THREE ITEMS]
\vspace{0.4em}
\noindent\textbf{Data Processing.}~The input data of this work consists of traffic trajectory data\footnote{Data comes from Didi Chuxing GAIA Initiative, https://gaia.didichuxing.com} and road network data\footnote{Data comes from OSMnx, https://www.openstreetmap.org/}. We first rasterize the urban area (\fig\ref{fig: overview}($a_1$)). Next, we aggregate the grids based on the number of road intersections (\fig\ref{fig: overview}($a_2$)). 

\vspace{0.4em}
\noindent\textbf{Prediction Interpretation.}~The prediction-Interpretation layer runs the traffic prediction model and provides interpretation to support the analysis tasks (\fig\ref{fig: overview}(B)). \trafps~applies the state-of-art traffic prediction model (\ie,  ST-ResNet~\cite{zhang2017deep}, CNN, DNN) to predict the urban traffic (\fig\ref{fig: overview}($b_1$)) and uses novel proposed Shapley-based measurements and newly designed visualization glyphs to provide interpretation(\fig\ref{fig: overview}($c_2$)).

\vspace{0.4em}
\noindent\textbf{Visual Explanation and Exploration of Predictions.}~To support interactively exploring the urban traffic prediction, we developed an interactive visual interface (\fig\ref{fig: overview}(C)) for~\trafps. The user interface of \trafps~is a web-based multi-view system, aiming to provide interactive visual analytics for domain experts on interpreting traffic flow prediction.

\vspace{0.4em}
%\noindent\textbf{System Architecture.}~
The data collection and preprocessing stages are executed offline on a GPU server with 6 GeForce GTX 1080Ti GPUs. \trafps~is developed and deployed on an Apple M1 Pro MacBook Pro with 16GB memory. The frontend of \trafps~is developed by Vue.js and D3.js and the backend of the system is supported by Flask. \fzz{As the computation process is an NP-hard problem, we use open source tool~\cite{lundberg2017unified} to approximate the Shapley value and get the interpretation tensor of the prediction results, which contains the \regionshap~of each grid in inflow and outflow matrix in 5 continuous timestamps that used to predict the next timestamp.}

% % % % % % % % % % % % % % % % % % % % % % % % % % % % % % % % % % % % % % % % % % % % % % % % % % % % % %% % % % % % % % % % % % % % % % % % % % % % % % % % % % % % % %
\section{Model}\label{sec: model}

In this section, we introduce the data processing phase in \sect\ref{subsec: data processing}, including geographical partition and regional aggregation. The details of the prediction-interpretation layer are described in \sect\ref{subsec: predict-interprete} where two key concepts, \regionshap~and \trajectoryshap~are introduced.

\subsection{Data Processing}\label{subsec: data processing}

The data processing layer supports necessary data preprocessing steps, which consist of two major steps: geographical partition and regional aggregation.

\vspace{0.4em}
\noindent\textbf{Geographical Partition.}~Due to the input data required for the selected traffic prediction model in our research, that should be a sequence of fixed-sized matrices. In addition, the results in~\cite{zhang2016dnn,zhang2019flow} show that the division can capture the latent spatial-temporal features within data. We first rasterize the area into grids. Specifically, the urban area is partitioned into an $M \times N$ grid map based
on the longitude and latitude ranges. We define the in-/out-flow as the measurement of trajectory data. Let $P$ represent the collection of trajectories. For a grid $(m,n)$ which locate at the $m^{th}$ row and the $n^{th}$ column, in-flow and out-flow at the timestamp $k$ are defined as:
\begin{equation*}
\whj{X^{in,m,n}_k = \sum_{t_k}|\{i>1|g_{i-1} \notin (m,n), g_{i} \in (m,n)\}|}
\label{equ:inflo}    
\end{equation*}
\begin{equation*}
\whj{X^{out,m,n}_k = \sum_{t_k}|\{i>1|g_{i} \in (m,n), g_{i+1} \notin (m,n)\}|}
\label{equ:outflow}    
\end{equation*}
where $t_k:g_1\rightarrow g_2\rightarrow ... \rightarrow g_{|{Tr}_k|}$ is one trajectory with $|t_k|$ sampling points, $k$ indicates the current timestamp, and $g_i \in (m,n)$ indicates the location of one sampling point $g_i$ located in the grid $(m,n)$. Therefore the in-/out-flow in the grid map $M \times N$ in the timestamp $k$ can be represented as $X_k \in \mathbb{R}^{2 \times M \times N}$. To better attribute each grid's contribution, and distinguish the in-/out-flow in each grid, we define the equivalent definition for \eq\eqref{equ:inflo} and \eq\eqref{equ:outflow}. Let $X$ be the flow matrix, we split each trajectory by the interval as a tensor. Let $\Omega$ be the set of
all trajectories, $T_i \in \Omega$ denotes a trajectory. $T_i^{in}$ and $T_i^{out}$ refer to a transfer presentation by the following constraint condition:
\begin{equation}
T^{in,i,j}_k = |\{k>1|g_{k-1}\notin (i,j)\wedge g_k \in (i,j)\}| 
\label{equ: constraint condition 1}
\end{equation}

and
\begin{equation}
T^{out,i,j}_k = |\{k\geq 1|g_{k}\in (i,j)\wedge g_{k+1} \notin (i,j)\}|
\label{equ: constraint condition 2}
\end{equation}

Intuitively, the trajectory flow tensor $T_k = [T_i^{in}, T_i^{out}]$ is equal to $X$ when only one trajectory considered in flow prediction. Therefore, $X^{in}$ and $X^{out}$ is defined as:
\begin{equation}
X^{in} = \sum_{T_k^{in}\in \Omega} T_k^{in}, X^{out} = \sum_{T_k^{out}\in \Omega} T_k^{out}
\label{equ:trajectory in/out flow}    
\end{equation}

Next, we count the passing trajectories for each grid to get the in-/out-flows. We will further discuss the relation of the size of the grids and the result in \sect\ref{sec: discussion}.

\vspace{0.4em}
\noindent\textbf{Regional Aggregation.}~In the urban environment, it is not a good choice to directly segment the urban area into grids or arbitrarily aggregate them into clusters. Such raster-like segmentation ignores the urban context. In addition, simply aggregating the grids into the same size may be inappropriate due to the unbalanced distribution of the urban traffic structure. Therefore, we use the number of intersections in the cluster, which is the key component of the urban structure, as the measurement of the regional aggregation phase. We apply K-means to aggregate the whole grids in the urban area into different clusters, which consider the number of intersections in the cluster. \fzz{In this research, we use the intersections' location (latitude and longitude) and the number of clusters (K value) as the input features. The termination condition for the algorithm is until no objects are reassigned to different clusters. Therefore, when the K-means algorithm stops, the distance from each intersection to the cluster's center is the shortest.} In \trafps, we map the traffic information on each intersection, and referring from \cite{feng2020topology}, we assume that the impact of each intersection is the same. In \trafps, we divide the selected region into 21 clusters aiming to achieve a good performance, more details on the number of clusters can be found in \sect\ref{sec: discussion}. 
% The result of the regional aggregation is shown in \fig\ref{fig: division}.

\subsection{Prediction-Interpretation Layer}\label{subsec: predict-interprete}
\subsubsection{Prediction Model}

The \trafps~is a post-hoc interpretation method using a well-trained model to predict the traffic first. We define our traffic prediction problem as given the historical observations $X_k$ for $k = 0, ..., t-1$, and predict $X_t$. In this research, according to the domain expert's suggestion, we apply ST-ResNet~\cite{zhang2017deep} in the \trafps. We divided the urban area into a $38 \times 36$ grid map and converted the trajectory data into in- and out-flow tensors with 10 minutes for each slice. The prediction problem uses the in- and out-flow of 5 nearest historical timestamps to predict the in-flow and out-flow tensor in the following timestamps.

\subsubsection{Shapley Value}\label{subsec: shapley value}
In the interpretation phase, we use a Shapley-based approach to interpret the traffic. The Shapley value is defined via a value function $val$ of players in $S$, and a feature value is its contribution to the payout, weighted and summed over all possible feature value combinations, thus the Shapley value $\phi_j(val)$ can be represented as:

\begin{equation}
\begin{split}
    &\sum_{S \in D}\frac{|S|!(p-|S|-1)!}{p!}(val(S\cup\{x_j\})-val(S)) \\
    &D = \{x_1,...,x_p \}\backslash \{x_j\}   
\end{split}
\label{equ:shapley}    
\end{equation}
where $S$ is a subset of the features used in the prediction model, $x$ the vector of feature values of the instance to be explained, and $p$ the number of features. $val_x(S)$ is the prediction for feature values in set $S$

\subsubsection{Region SHAP}\label{sec: region shap}

\begin{table}[t]
	%\scriptsize
	\centering
	\renewcommand\arraystretch{1.0}
	\caption{\whj{Partial Symbols Description.}}
	\label{tab:symbol}
	\begin{tabular}{c|l}
		\toprule
		Notation & Description\\
		\hline
		$S$ & subset of the features\\
             $T$ & trajectory tensor\\
             $t$ & single trajectory \\
             $X$ & flow tensor \\
             $C$ & subset trajectory tensor\\
             $\phi$ & value function\\ 
		%\setminus
		\bottomrule
	\end{tabular}
\end{table}

The original assumption of the Shapley Value is additive feature attribution~\cite{lundberg2017unified}, meaning that the total benefit could be equal to the contribution summation for each component. This is consistent with the~\eq\eqref{equ:trajectory in/out flow} that the flow matrix $G$ is composed of the trajectory flow tensor, both in model input and output. Therefore, we propose the \regionshap, which represents the regional contribution derived from the traffic flow (trajectories). Given the input:
\begin{equation}
\whj{X_t = [X^{in}, X^{out}] \in \mathbb{R}^{2\times M\times N}}
\label{equ: input xt}    
\end{equation}
and the model output:
\begin{equation}
\whj{Y_{t+1} = [Y_{t+1}^{in}, Y_{t+1}^{out}] \in \mathbb{R}^{2\times M\times N}}
\label{equ: output yt}    
\end{equation}
    Where $M$ and $N$ denote the height and the width of flow matrix $X$, therefore, we define the regional Shapley value named \regionshap~that shows the effect $gird(i,j)$ has on the prediction of the $grid(u,v)$ is:
\begin{equation}
    \phi(Y_{t+1}^{c_1,u,v} \rightarrow X_{t}^{c_2,i,j})
    \label{equ: region SHAP}    
\end{equation}
where $c_1$ and $c_2$ denote the channels: in-/out-flow. Intuitively, \regionshap~is region to region based explanation. We use an open-source tool~\cite{lundberg2017unified} to compute the Shapley value and get the interpretation tensor of the prediction results, which contains the Shapley value of each grid in in-flow and out-flow tensor in $k$ different timestamps that are used to predict the next timestamps.

\subsubsection{Trajectory SHAP}\label{sec: trajectory shap}
In our problem, \fzz{we make an analogy that the trajectories $t$ in \trajectoryshap~can be regarded as the players in Shapley Value in \sect~\ref{subsec: shapley value}}. Therefore, we have the theorem with 
\begin{theorem}\label{the:shap_proof}
	Given a region explanation saliency maps $E(g^{c,i,j}_t)$, 
	the contribution of each trajectory in $g^{i, j}$ is $E(g^{c,i,j}_t \rightarrow t_r) =\frac{1}{\mid C^{c,i,j} \mid}  E(g^{c,i,j}_t) $,  where the subset trajectories $C$ is defined as 
	\begin{align*}
		C^{in,i,j} &=\left\{ t_r \in C^{in,i,j} \mid (i,j)\in \{  k>1 \mid v_k\} \right\} \\
		C^{out,i,j} &=\left\{ t_r \in C^{out,i,j} \mid (i,j)\in \{ k< N \mid v_k\} \right\}
	\end{align*}
	where $N$ is the length of trajectory $t_r$, and $C^{in,i,j}$ and $C^{out,i,j}$ is to collect the trajectories where flow in or flow out the region $g^{i, j}$ in $t$ time slot respectively. 
	
	\textbf{proof}: Given the region explanation saliency maps $E(g^{c,i,j}_t)$, the goal is to explain the contribution $g^{i, j}$ is $E(g^{c,i,j}_t \rightarrow t_r)$ of each trajectory $t_r \in C^{c,i,j}$. The contribution of each trajectory with the Shapley Values format is 
	\begin{equation}
	    \begin{split}
\whj{  \phi_{j}(v a l)=\sum_{S} \frac{|S| !(p-|S|-1) !}{p !}\left(val\left(S \cup\left\{x_{j}\right\}\right)-val(S)\right), \nonumber}
		\end{split}
	\end{equation}
	where $n= \mid C^{c,i,j} \mid$. For each pairwise trajectory $(t_i, t_j) \in C^{c,i,j}$, due to the definitions ~\eq\eqref{equ:inflo} and ~\eq\eqref{equ:outflow} do not distinguish the trajectories, we have 
	\begin{align}
\whj{val(S \cup\left\{t_{n} \right\}) = val(S \cup\left\{t_{m} \right\}) , \quad \forall S \in C^{c,i,j}.}
	\end{align}
 Therefore, each trajectory contribute $\phi(t_m)$ is equal and $\phi(t_m)=E(g^{c,i,j}_t \rightarrow t_m) =\frac{1}{\mid C^{c,i,j} \mid}  E(g^{c,i,j}_t)$.  
\end{theorem}
According to Theorem \ref{the:shap_proof}, the trajectory  relevant score at channel $c$ can be defined as 
\begin{align}
	E(t^c_r)=\sum_{(i,j) \in v_a \wedge v_a \in t_r} E(g^{c,i,j}_t \rightarrow t_r),
\end{align}
where, $t_r : v_{1} \rightarrow v_{2} \rightarrow \cdots \rightarrow v_{|t r|}$ is a single trajectory. Noted that, when $c$ stands for out-flow, the previous path within $0$ to $|t r|-1$ will be calculated, on the contrary, $c$ presents in-flow, 1 to $|t r|$ will be counted. Intuitively, it is the result of the summation of all paths where $t_r$ go through. 

In our work, to interpret the traffic prediction in detail, we propose the \trajectoryshap, which is the region to trajectory explanation. When given a \regionshap~value $\phi(Y_{t+1}^{c_1,u,v} \rightarrow X_{t}^{c_2,i,j})$ where $x_t^{in,i,j}>0$ and the sub-trajectories are defined as:

\begin{equation}
    \whj{C^{in,i,j} = \{t_r \in (I,J)| (i,j) \in \{k > 1 | \quad g_k\}\}}
    \label{equ: sub-trajectory in}    
\end{equation}
and 
\begin{equation}
    \whj{C^{out,i,j} = \{t_r \in (I,J )| (i,j) \in \{k > N | \quad g_k\}\}.}
    \label{equ: sub-trajectory out}    
\end{equation}
The contribution of each trajectory in $x^{i,j}$ which is the \trajectoryshap~ $\phi(Y_{t+1}^{c_1,u,v}\rightarrow T_{r}^{c_2})$ can be represented as:
\begin{equation*}
    \phi(Y_{t+1}^{c_1,u,v}\rightarrow T_{r}^{c_2})= \sum_{i,j} \frac{1}{|C^{c_2,i,j}|}\phi(Y_{t+1}^{c_1,u,v} \rightarrow X_{t}^{c_2,i,j})
    \label{equ: trajectory SHAP}    
\end{equation*}
Where $N$ is the length of trajectory $t_r$, and $C^{in,i,j}$ and $C^{out,i,j}$ is to collect the trajectories where flow in or flow out the region $g^{i,j}$ in $t$ time slot respectively.

% % % % % % % % % % % % % % % % % % % % % % % % % % % % % % % % % % % % % % % % % % % % % % % % % % % % % % % % % % % % % % % % % % % % % % % % % % % % % % % % % % % % % % % % % % % % % % % % % % % % % % % % % % % % % % % % % % % % % % % % % % % % % % % % % % % % % % % % % % % % % % % % % % % 

\section{Visual Design}\label{sec: visual design}

This section first discusses the design rationales of \trafps, which provides the design tasks for \trafps, then introduces three interactive and interpretation-assisted visualization views in \trafps~that ease and improve analysts' analysis process of understanding the prediction.

\subsection{Design Rationale}\label{subsec: design rationale}

According to the nested model for visualization design and validation~\cite{munzner2009nested}, we summarize a three-stage visual analytical workflow and specify particular design tasks in each stage for \trafps~with the domain experts. The three stages are summarized from coarse grain granularity to fine grain granularity. Each designed view corresponds to the corresponding stage.

\begin{figure}[htbp]
\includegraphics[width=1\linewidth]{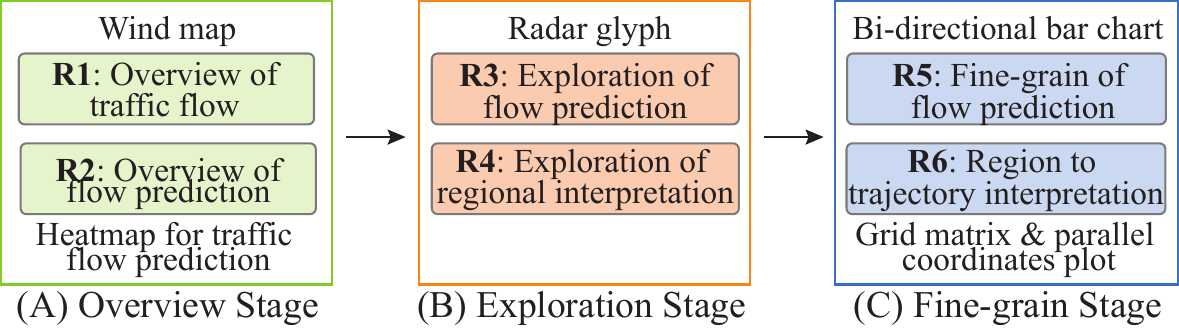}
   \caption{\fzz{The design rationale consists of three stages, including (A) overview stage, (B) exploration stage, and (C) fine-grain stage.}}
\label{fig: design rationale}
\end{figure}
\vspace{0.4em}

\noindent\textbf{Overview Stage~(\fig\ref{fig: design rationale}(A)):}~This stage provides an overview of traffic flow, and visualizes traffic flow change in a coarse granularity. 

\textbf{R.1: Overview of traffic flow.} Users can observe a global or local tendency of traffic flow, including the volume of the traffic, and the average speed of the primary road.

\textbf{R.2: Overview of flow prediction.} Users can see the prediction result globally and observe the relative changes. Furthermore, this task enables the user to distinguish the current condition and the prediction result intuitively.

\vspace{0.4em}
\noindent\textbf{Exploration Stage~(\fig\ref{fig: design rationale}(B)):}~This stage supports users exploring traffic flow prediction and its interpretation for a specific region in the urban area. The urban area is segmented into regions (\sect\ref{subsec: data processing}(Regional Aggregation), and we use the \regionshap~(\sect\ref{sec: region shap}) to provide region-to-region interpretation.

\textbf{R.3: Exploration of flow prediction.}~Users can observe the traffic flow prediction in a specific region at a certain period and can be aware of the tendency of the traffic condition in the future (rise/down). In addition, users can also compare the prediction between different regions. 

\textbf{R.4: Exploration of regional interpretation.} Users can get the region-to-region interpretations according to the \regionshap. The design can also combine with the prediction result to increase space utilization.

\vspace{0.4em}
\noindent\textbf{Fine-grain Stage~(\fig\ref{fig: design rationale}(C)):}~This stage supports users in delving into one region and analyzing the traffic flow at a fine-grained level. A region to trajectory interpretation derived from \regionshap~is provided for more detailed and direct interpretation.

% \textbf{G1: Division of Grid}. Grids should be appropriate in size and easy to be manipulated 

\textbf{R.5: Fine-grain of flow prediction.}~Users can be provided with a more detailed division for observing the flow prediction in each region. The detailed information should be highly correlated with the spatial dimension.
% as number of grids can be much bigger than cluster.

\textbf{R.6: Region to trajectory interpretation.}~Users can analyze the traffic flow according to the \trajectoryshap, which is proposed to provide a region to trajectory interpretation. Users can look up the most basic information to support decisions.

\subsection{Map-Trajectory View}

\begin{figure}[htbp]
\includegraphics[width=0.95\linewidth]{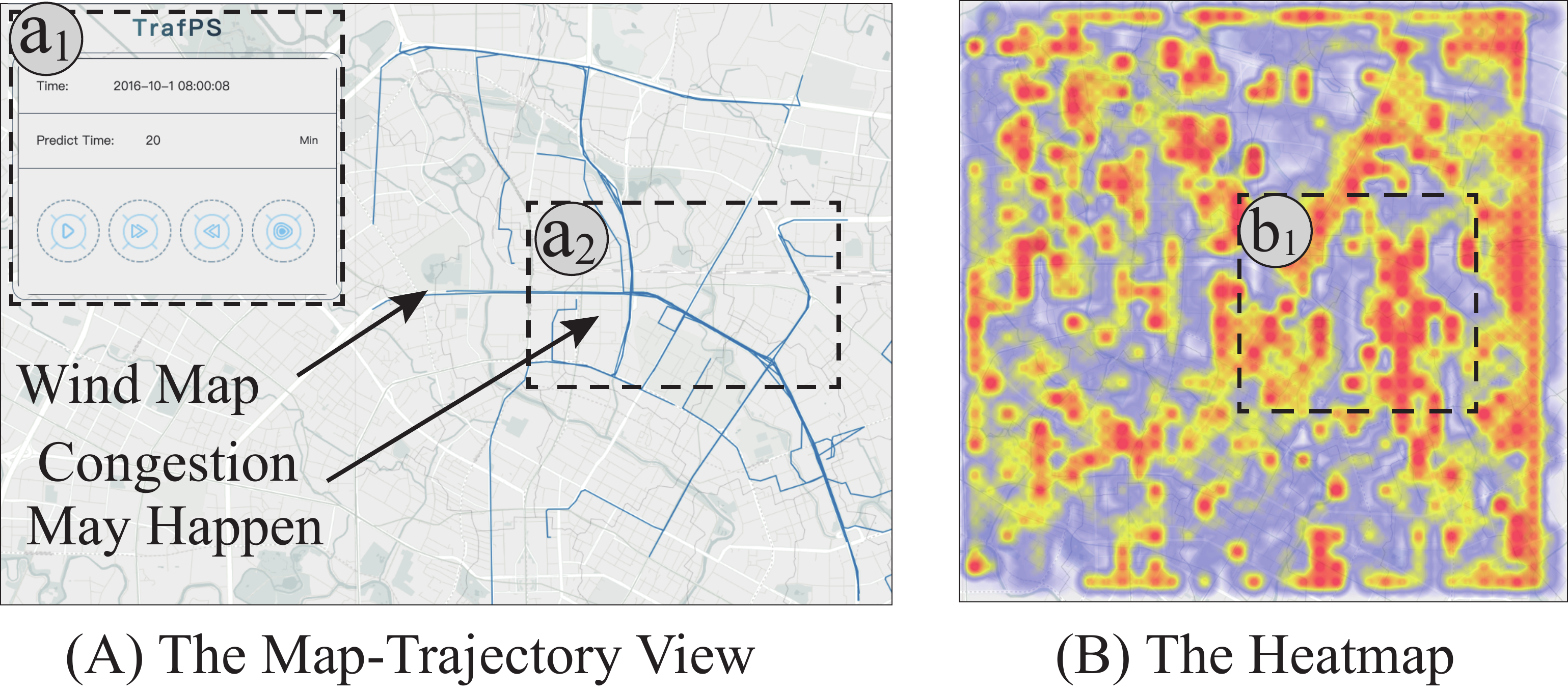}

   \caption{The design of map-trajectory view. The map-trajectory view consists of three parts, including ($a_1$) a dashboard, ($a_2$) a geography map, and (B) a traffic flow prediction heatmap.}
\label{fig: case1-mapview}
\end{figure}

The map-trajectory view aims to provide analysts with an overview of the current traffic situation (\textbf{R.1}) and the traffic prediction (\textbf{R.2}) in the urban area through visualization techniques. The map-trajectory view consists of three parts, including a dashboard (\fig\ref{fig: case1-mapview}($a_1$)), a geography map (\fig\ref{fig: case1-mapview}($a_2$)), and a traffic flow prediction heatmap (\fig\ref{fig: case1-mapview}(B)). 

The dashboard provides several operators to support the analysis process. Analysts can control the analytical process from the temporal dimension. Analysts can pause the analytical process at any timestamp to analyze traffic flow in detail or switch to locate specific timestamps when the \trafps~runs. In addition, at the bottom of the dashboard, we leave a space to locate the radar glyph of the selected area. When all settings are ready, the trajectory data will be dynamically visualized on the geography map as a simplified wind map (\fig\ref{fig: case1-mapview}(A))\cite{shi2020urbanmotion}. In the wind map, we use the shades to encode the current traffic volume, the more obvious the shades, the higher volume of the traffic flow. We also use the frequency of the wind map to encode the average road speed, that is, the higher the frequency of the wind map, the higher the average speed of the road, which means that the traffic condition is good, and vice versa.

\begin{figure}[htbp]
\includegraphics[width=1\linewidth]{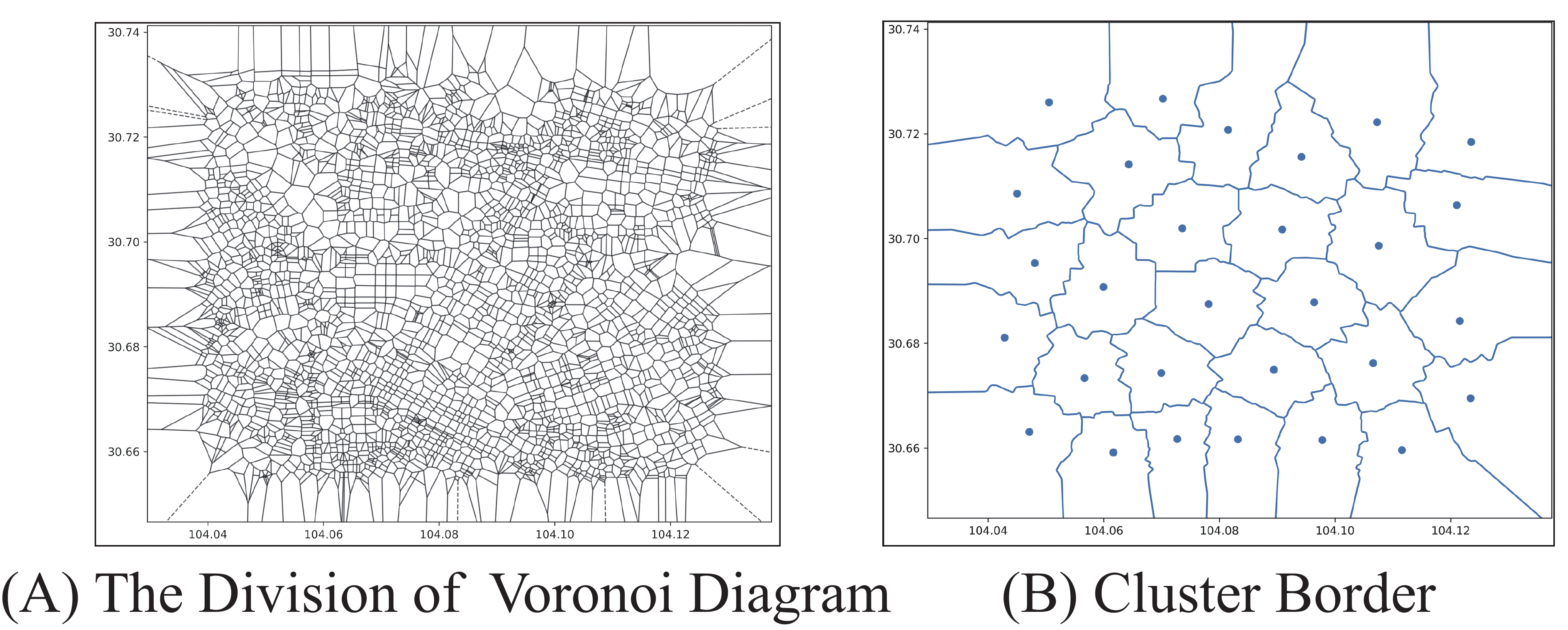}
   \caption{The results of regional aggregation. (A) is the result of the division of the Voronoi Diagram for the urban area, and (B) is the visual effect of the border of each cluster.}
\label{fig: division}
\end{figure}

\begin{figure}[htbp]
\includegraphics[width=1\linewidth]{Radar-Glyph.pdf}
% \begin{center}
%    %\includegraphics[width=0.8\linewidth]{egfigure.eps}
% \end{center}
   \caption{The (A) design and the (B) example of radar glyph.}
\label{fig: radar glyph}
\end{figure}
We re-organize the border of each aggregated cluster mentioned in the \sect\ref{sec: model}(Regional Aggregation) due to the experts' suggestions and the real situation aspects. During the interview with the domain experts, they mentioned that using the grids' border as the cluster is not a good choice. In the division of the grid, we divide the urban area into same-size grids according to the longitude and latitude, ignoring the distribution of the intersections in real-world situations. Therefore, we employ the Voronoi diagram to generate each intersection with a polygon containing only the intersection itself, and every point in a given polygon is closer to its generating intersections than to any other. Compared with K-means, Voronoi also utilizes the intersection as the input, but generates a more reasonable border based on the density of intersections at a specific area.  With the help of the Voronoi diagram, we get the border of each intersection and show it on \fig\ref{fig: division}(A). We merge the border of each intersection in the same cluster, produced by K-means, and replace the grid's border distributed at the edge of each cluster with the border of the intersection (\fig\ref{fig: division}(B)).

Moreover, we apply a heatmap to distinguish the predictive traffic condition from the current traffic flow by presenting data in high comparability (\fig\ref{fig: case1-mapview}($b_1$)), where darker color means more heavy traffic flow. When the analysts input the prediction period $t$, and then the predicted traffic condition will be visualized as the heatmap.

\subsection{Radar Glyph View}

\begin{figure*}[htbp]
\includegraphics[width=0.95\linewidth]{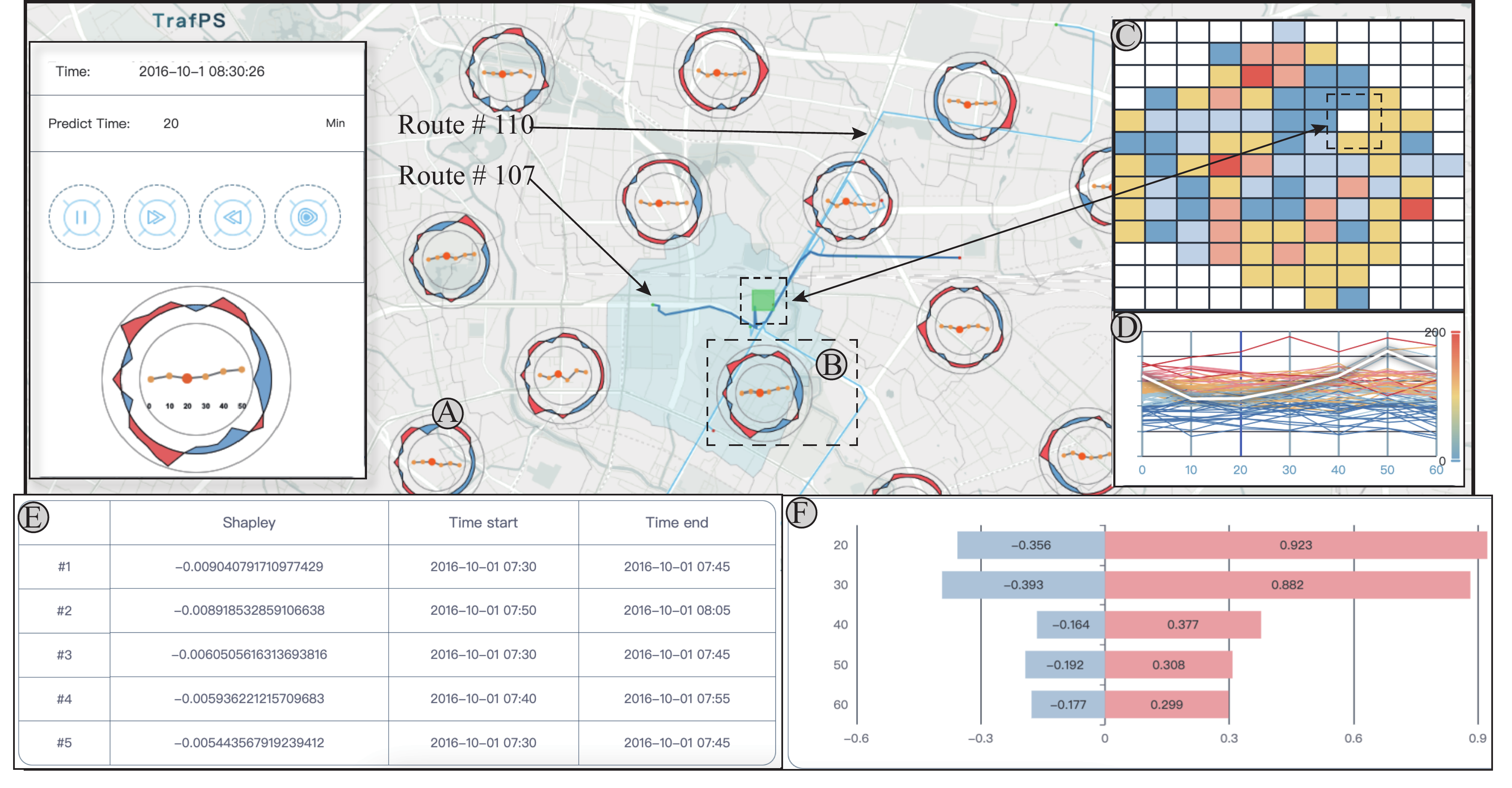}
   \caption{The the user interface of \trafps. The \trafps~consists of three parts, including map-trajectory view, (A) \& (B) radar glyph view, and fine-grained grid view. The components of the fine-grained grid view include (C) a grid matrix, (D) a parallel coordinates plot, (E) a detailed information table, and (F) a bi-directional bar chart.}
\label{fig: case1-big}
\end{figure*}

The newly designed radar glyph aims to represent the traffic flow prediction and provide the interpretation at a medium level (\textbf{R.3} and \textbf{R.4}). In this research, we encode prediction and interpretation information into one glyph simultaneously to effectively make the situation-aware. Radar glyph placed at the centroid of each segmented region to summarize the traffic flow prediction of the region and its corresponding interpretation (\fig\ref{fig: radar glyph}(A)). Clicking the selected region can magnify the glyphs and gain a more explicit representation for detailed exploration of the insights (\fig\ref{fig: radar glyph}(A)). The radar glyph consists of two parts: the inner prediction and the outer interpretation.

% \vspace{0.4em}
\noindent\textbf{Traffic Flow Prediction:}~In the inner part of the radar glyph, we use a line chart placed in the center of the glyph to represent the size of the predicted traffic flow. The dot points are arranged in sequence along the line chart. Dot points far away from the timeline represent a significant predictive value of the traffic flow at the corresponding future period $t$. The highlighted dot point represents the selected period for prediction (\eg, in \fig\ref{fig: radar glyph}(A) the highlighted point.). For example, according to the prediction results, \fig\ref{fig: radar glyph}(A) shows that the traffic flow will increase in the following time as the dot points in the future are located further away from the timeline than the points representing the current time, especially in 30 minutes later the traffic flow will reach a high level. Furthermore, the dot point represents the period in 20 minutes is highlighted, which means that the user selects 20 minutes as the future period to interpret.

% \vspace{0.4em}

\noindent\textbf{Traffic Flow Interpretation:}~In the outer part of the radar glyph, we use the distribution of two diverging polygons to represent the correlation between the segmented region’s neighbor area and the segmented region’s traffic flow prediction at the selected future period $t$ by the geographical directions of these surrounding areas, the bigger the polygon, the more influential the positive (\textcolor{red}{red}) or negative (\textcolor{blue}{blue}) correlation. For instance, the user selects the future period $t$ as 20 minutes which shows that the outer glyph will focus on the interpretation of the next 20 minutes. In \fig\ref{fig: radar glyph}(A), we found that the surrounding area of the polygon in both the west and the east direction are large areas in size, and both two directions are colored red, contributing to the increase in traffic flow. As a result, experts can restrain the traffic flow or adjust the traffic network in those areas tentatively to avoid potential traffic jams which may occur in the future 20 minutes. Furthermore, users can find the key region/road that may affect future traffic flow by analyzing a series of adjacent radar glyphs. For instance, from the interpretation parts of three adjacent radar glyphs in \fig\ref{fig: radar glyph}(B), we can find that in the three glyphs commonly related regions, all the polygon areas in the three glyphs are significant and colored red, which indicate that this region may play an essential role in impacting local traffic flow. For detailed searching this region, we found Road \#121, which may affect the traffic according to the radar glyph.

\subsection{Fine-grained Grid View}

The fine-grained grid view aims to provide a representation of the traffic flow prediction and its interpretations (\textbf{R.5} and \textbf{R.6}) on a fine-grained level, \ie, the specific regions and the key intersections. To support the requirements, we divide each segmented region into grids. Then we provide a fine-grained grid-based traffic flow analysis through several components, including a grid matrix (\fig\ref{fig: case1-big}(C)), a parallel coordinates plot (\fig\ref{fig: case1-big}(D)), and a bi-directional bar chart (\fig\ref{fig: case1-big}(F)).

The grid matrix (\fig\ref{fig: case1-big}(C)) reflects the current traffic condition for the selected segmented region. The relative positions of the grids in the grid matrix are consistent with the grids on the geography map and correspond to each other (\eg, in \fig\ref{fig: case1-big} the highlighted region and the region in \fig\ref{fig: case1-big}(C)). Moreover, we use color to encode the volume of the flow, the large to the small flows correspond to the red to yellow to blue. Clicking the grid in the grid matrix will highlight the corresponding time series in the parallel coordinates plot (\fig\ref{fig: case1-big}(D)).

The design of the parallel coordinates plot (\fig\ref{fig: case1-big}(D)) is with the same color scheme in the grid matrix, reflecting the change of the prediction for each grid, aiming to accurately visualize traffic flow trends in the corresponding grid.

In order to analyze the traffic flow in some specific places, in \trafps, users can select one grid in the grid matrix to analyze~(highlighted in (\fig\ref{fig: case1-big}), and the top K trajectories (in the case study, we set K as a default number 5) that have the most correlation or impact on the traffic flow prediction of this grid will appear on the map. The trajectory's color means the nature of correlation, as same as the color in the radar glyph.  A table list contains the details of the top K trajectories~(\fig\ref{fig: case1-big}(E)), and a bi-directional bar chart aggregates the correlation of each trajectory into different time channels~(\fig\ref{fig: case1-big}(F)).

% % % % % % % % % % % % % % % % % % % % % % % % % % % % % % % % % % % % % % % % % % % % % % % % % % % % % % % % % % % % % % % % % % % % % % % % % % % % % % % % % % % % % % % % % % % % % % % % % % % % % % % % % % % % % % % % % % % % % % % % % % % % % % % % % % % % % % % % % % % % % % % % % % % % % % % % % % % % % % % % % % % % % % % % % % % % % %

\section{Evaluation}\label{sec: evaluation}

The \trafps~aims to provide an effective and intuitive interpretation to help understand models’ predictions and support decisions to the domain experts. This section conducts two case studies that the \trafps~satisfied the requirements of identifying key routes and supporting decision-making. The summary of the feedback from an expert interview to demonstrate the feasibility, usability, and effectiveness is presented at the end of this section.

\vspace{0.4em}
\noindent\textbf{Data preparation and parameter configuration:}~We collaborate with the domain experts (\textbf{E.A} and \textbf{E.B}) and perform the case studies on a real-world dataset that contains two-month taxi trajectory data (from 10/01/2016 to 11/31/2016) in Chengdu, China. Road network data contains the traffic intersection data in the exact location of Chengdu. In the two case studies, we chose 21 as the number of the aggregated region.
% as we want more aggregated regions to provide more local and detailed information after trying 0 to 27. 
Moreover, in the case studies, we set the future time to conduct the corresponding interpretation as 20 minutes.

\subsection{Identify Key Routes: Which Contribute the Most}
We first demonstrate how the \trafps~can reveal the traffic congestion in the near future and help analysts' decision-making by identifying the key routes.

\vspace{0.4em}
\noindent\textbf{Procedure:}~The analyst first operates the dashboard (\fig\ref{fig: case1-mapview}($a_1$)) to select which time to analyze. He selects an early morning (8:30 am) on a public holiday (10/01/2016, the National Day of China) as the research time. After setting the time, the map-trajectory view generates all real-time data moving trajectories (\fig\ref{fig: case1-mapview}(A)) the frequency of the wind map indicates the average speed of each route, the direction of the wind map shows the main tendency of the traffic flow and the thickness of the routes reflects the density of the vehicle on the road. From the map-trajectory view, the traffic flow tends to move towards the center of the map, one of the main roads connecting the urban area and the suburban area, indicating traffic congestion may happen. In addition, the thickness wind map around the city (\fig\ref{fig: case1-mapview}($a_2$)) also shows a large number of traffic flows that may lead to congestion. Then he projects a heatmap (\fig\ref{fig: case1-mapview}(B)) to see traffic flow prediction 20 minutes later. The heatmap reveals that several areas with the darkest color will have congestion, \eg, the center of the map (\fig\ref{fig: case1-mapview}($b_1$)).

Being aware that the center of the map may have traffic congestion in 20 minutes, the analyst observes the radar glyphs in the selected region to see which neighboring areas contribute to future congestion. 
% As shown in \fzz{figure xxx}, segmented regions \#1 and \#2 are located in the center of the map. 
According to the line charts in the radar glyphs (\fig\ref{fig: case1-big}(A)), it is intuitive that the traffic flow will increase 20 minutes later. And the inner polygons reveal that the neighboring areas from the north have a more effective correlation. Therefore, the traffic management department can temporarily prevent heavy congestion by constraining the traffic flow in these areas, and the citizens can choose to make a detour to avoid the coming traffic congestion on such routes.

The analyst takes further steps and handles the latent traffic congestion more efficiently and precisely through visualization and interpretation in the fine-grained grid view (\fig\ref{fig: case1-big}(C), \fig\ref{fig: case1-big}(D), \fig\ref{fig: case1-big}(E) and \fig\ref{fig: case1-big}(F)). From the parallel coordinates plot, he finds some yellow lines will get a sharp increase (\fig\ref{fig: case1-big}(D)), which may be a prominent location for traffic congestion. The analyst then clicks the corresponding grids in the grid matrix (\fig\ref{fig: case1-big}(C)) to project the top five significant routes. The red routes mean they all promote traffic congestion. The analyst realizes appropriate reactions should be made on route \#107 and route \#110 (\fig\ref{fig: case1-big}) to prevent traffic congestion. Moreover, the analyst checks the detailed information table (\fig\ref{fig: case1-big}(E)) and the bi-directional bar chart (\fig\ref{fig: case1-big}(F)), which list the top five routes that may affect traffic. From the bi-directional bar chart, it is obvious that the traffic that happened 20-30 minutes ago affects the prediction most. This phenomenon indicates that the analyst needs to pay more attention to the traffic in that time slot.

\vspace{0.4em}
\noindent\textbf{Findings:}~The analyst found that the \trafps~can provide different predictions and corresponding interpretations between three visualization views from the case study. By interacting with the \trafps~the analyst is encouraged to customize the analysis workflow.

\subsection{Support Decisions: How to Improve Urban Traffic}

This case aims to show how our method can help urban planners on planning the traffic infrastructure in the urban area by revealing the trip law.

\begin{figure}[htbp]
\includegraphics[width=1\linewidth]{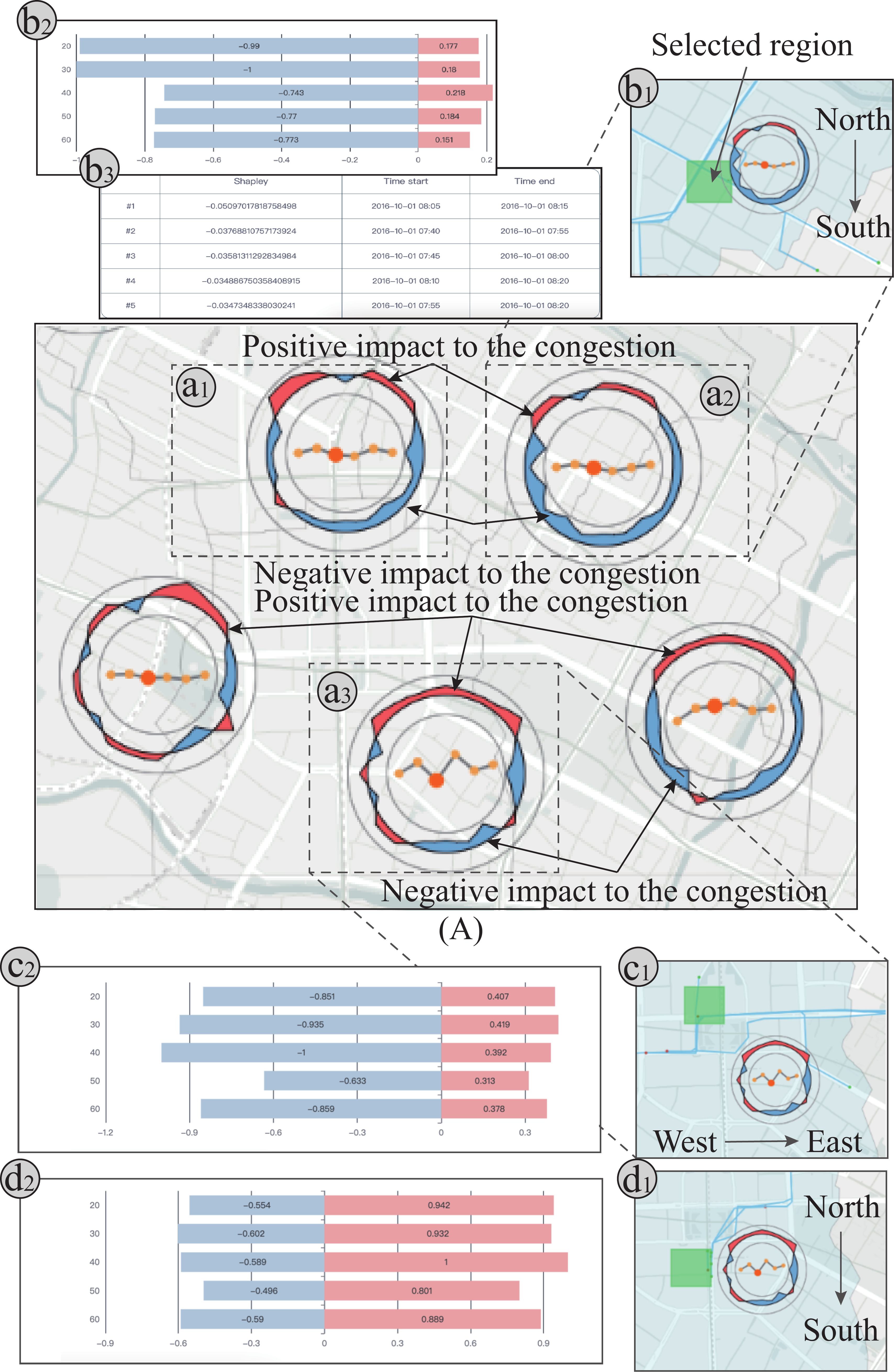}

   \caption{The \trafps~can support decision-making in the planning of reversible lanes. (A) The radar-glyphs can reveal the directions of the traffic impact on the selected region, and ($b_2$,$b_3$,$c_2$, and $d_2$) the detailed information shows whether the selected regions contribute a positive or negative impact on traffic congestion.}
\label{fig: case2}
\end{figure}

\vspace{0.4em}
\noindent\textbf{Procedure:}~The analyst chooses the working day morning as the research timestamp and sets 20 minutes as the predicted time. After the configuration, he selects the city center as the AoI. The analyst found that the radar glyphs distributed near the center of the city look different (\fig\ref{fig: case2}(A)), as in the inner part of these glyphs, it is intuitive that most of the contributions to the traffic congestion in these five areas come from the north. In addition, in the glyphs \fig\ref{fig: case2}($a_1$) and \fig\ref{fig: case2}($a_2$), it is obvious that the south side contributes more to decreasing the impact of traffic congestion.

The analyst clicks the glyph (\fig\ref{fig: case2}($a_2$)) to explore this phenomenon further and selects one grid that reveals the most traffic flow. The parallel plot shows in the selected grid, that the volume of the traffic may remain at a high level in the following 60 minutes. The detailed information is shown in \fig\ref{fig: case2}($b_3$). From the bi-directional bar chart~(\fig\ref{fig: case2}($b_2$)), it is obvious that all the trajectories in this grid negatively affect the impact of traffic congestion in this region. By checking the detailed information~\fig\ref{fig: case2}($b_3$), we found that the top five routes with the most significant impact can be classified from the north to the south.

Due to most of the key routes associated with the selected grid coming from region ~\fig\ref{fig: case2}($a_2$), the analyst then clicks region~\fig\ref{fig: case2}($a_3$) for further exploration. From the radar glyph (\fig\ref{fig: case2}($a_3$)), he found that the contribution of the traffic in the next 20 minutes is more comes from the north. To prove this phenomenon, he selects the two grids (\ie, \fig\ref{fig: case2}($c_1$) and \fig\ref{fig: case2}($d_1$)) that many routes come from the north and with high traffic volume, respectively. From the bi-directional bar chart (\fig\ref{fig: case2}($c_2$)), we can find that in the grid (\fig\ref{fig: case2}($c_1$)) shows a negative impact on traffic congestion, the most significant routes to affect the traffic congestion come from the west and may go through to the east. He then checks the grid (\fig\ref{fig: case2}($d_1$)) in this region and found that the tendency in the bi-directional bar chart shows a positive impact (\fig\ref{fig: case2}($d_2$)) on the traffic congestion and shows most of the routes reveal a north to south direction.

\vspace{0.4em}
\noindent\textbf{Findings:}~From the manipulation, the analyst found that most of the key routes affecting the selected grids' traffic congestion have the same direction, reflecting traffic patterns in peak hours (\textbf{E.A}). Therefore, \trafps~can assist urban planners in making decisions on the planning of reversible lanes.

\subsection{Expert Interview}\label{sec: expert interview}

We interviewed five domain experts to evaluate our research on feasibility, usability, and effectiveness. The first expert (\textbf{E.A}) is a research assistant professor at the University T, and his research interest lies in urban computing. The second expert (\textbf{E.B}) is an analyst at a consulting firm with a degree in urban planning. In addition, both the experts (\textbf{E.A} and \textbf{E.B}) are our collaborators, and \textbf{E.A} is also our co-author for this paper. Furthermore, we also sent invitations to five researchers who have publications in the field of urban-related top journals or conferences in recent five years. Three of them (\textbf{E.C}, \textbf{E.D}, and \textbf{E.E}) accept the invitation, and all three researchers have Ph.D. degrees. Before the interview with the five experts, we briefly introduced our work and played a video of the \trafps~with the case studies. All experts provided valuable feedback, and \textbf{E.C}, \textbf{E.D}, and \textbf{E.E} gave helpful suggestions based on their research backgrounds. We summarize the interview as follows.

\vspace{0.4em}
\noindent\textbf{Feasibility \& Usability.}~The experts all agreed that introducing the Shapley value to the urban traffic interpretation is a good attempt, and the design of the visual interface is intuitive to traffic interpretation, especially in the interpretation of traffic congestion prediction and helpful to decision-making. \textbf{E.A}, \textbf{E.C}, and \textbf{E.D} mentioned that it is good to apply Shapley value to the current research, for Shapley value is a mature method in interpretation and has been proven to be useful and can reveal the contributions from every input feature. Furthermore, all the interviewed experts claimed that the visual analytics tool is easy to use and can facilitate the decision-making process in urban planning. \textbf{E.B} mentioned that this interactive visual analytics approach is important to his work. When summarizing the analysis result to the leaders for approval, it is challenging to convince the seniors without sufficient domain knowledge. Therefore, this work can help to understand the model result and make further decisions through the step-by-step operation.

\vspace{0.4em}
\noindent\textbf{Effectiveness.}~All the experts agreed that the \trafps~could interpret the urban traffic, which may be effective for urban-related work. \textbf{E.B} shared his work experience in urban emergency management. When an emergency happens, he has to distinguish the key route for the emergency response department and help to plan the route in advance. However, using the current prediction model, he can only find which road may have congestion. However, he cannot acquire where the congestion comes from or which region may affect the traffic condition in the selected area most. \textbf{E.B} said that the \trafps~could improve the interpretation of his work, and more importantly, it can reveal the contribution to the traffic from the neighbor of the selected region.

\vspace{0.4em}
\noindent\textbf{Suggestions.}~According to the interview with the domain experts, they provide fruitful suggestions and advice on a future direction based on their background. \textbf{E.D} and \textbf{E.E} said that they are interested in the second case study, in which \trafps~can support decision-making. Both of them suggest that providing a period range for managing the traffic will be better. For example, they hope that \trafps~can provide experts with a specific period for controlling the reversible lane. They also said that it is indeed a challenging task for this function. \textbf{E.B} mentioned that using prediction results to plan the urban infrastructure is not a good choice for the urban planning task. The existing work proved that using historical data to mine the pattern performs well. Nevertheless, using the prediction model to predict the traffic and make a decision in a short time (\ie, emergency response management) is reasonable. So \trafps~is a good method to control the reversible lane and traffic signal when an emergency happens. \textbf{E.C} mentioned that if we want to deploy the \trafps~in the real environment, we can also explore the real-time traffic data (\eg, loop sensor data, DSRC data, VD data and etc.) of the urban traffic, which will make a more comprehensive analysis of the traffic.

% % % % % % % % % % % % % % % % % % % % % % % % % % % % % % % % % % % % % % % % % % % % % % % % % % % % % % % % % % % % % % % % % % % % % % % % % % % % % % % % % % % % % % % % % % % % % % % % % % % % % % % % % % % % % % % % % % % % % % % % % % % % % % % % % % % % % % % % % % % % % % %
\section{Discussion}\label{sec: discussion}

\noindent\textbf{Number of Aggregated Clusters.}~In \trafps, to facilitate the users' understanding of the context of the urban environment, we aggregate the region into different clusters, not just using the grid (the single unit for prediction). We apply the K-means algorithm to cluster the urban area and aim to find a suitable number of clusters. Due to the importance of the types of roads (\eg, freeway, primary, secondary), the capacity of each road is different. Therefore, as~\cite{feng2020topology} mentioned that the urban intersections, to some extent, can reflect the traffic load, we use the number of the intersections to reflect the traffic load, which assumes that the traffic load for different roads remains the same. For each cluster, we aim to keep a balance on the traffic load, we keep the number of intersections in each cluster at the same level and use the variance to measure the stability under different cluster numbers:

\begin{figure}[htbp]
\includegraphics[width=1\linewidth]{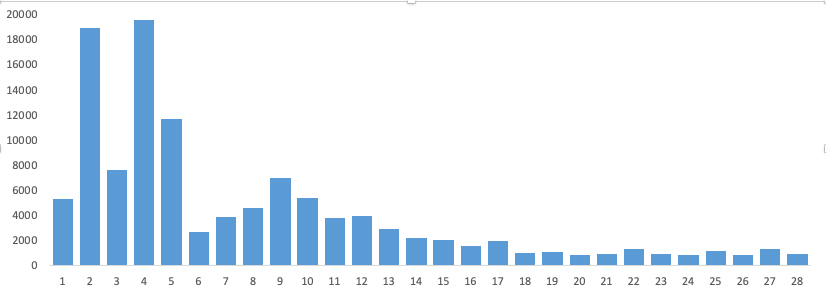}
% \begin{center}
%    %\includegraphics[width=0.8\linewidth]{egfigure.eps}
% \end{center}
   \caption{The statistical information of parameter selection of K-means algorithm.}
\label{fig: k-means}
\end{figure}

\begin{equation}
    \sigma^2 = \frac{\sum_{I_i \in I} {(\Bar{I} - I_i)}^2}{N}
    \label{equ: discussion}    
\end{equation}
where $I_i$ is the number of the intersection in each cluster, $I$ is the set of all the clusters, $\Bar{I}$ is the average of the intersection number, and $N$ is the number of the cluster. \fig\ref{fig: k-means} is the result, in which we set the number of clusters in the selected region from 1 to 26. When the number of clusters is small, the variance of the intersection in each cluster is large, but when the number of clusters increases, the variance becomes significantly smaller and remains at a stable level. Therefore, we arbitrarily choose 21 from the numbers greater than 17 as the $K$ value of K-means in this study.
%tab1
\begin{table*}[h!t]
\center
\caption{The statistical information of the key routes predicted by different prediction models.}
\begin{tabular}{lllllll}
\toprule
\multirow{2}{*}{\diagbox{Top K Routes}{Models}} & \multicolumn{3}{c}{CNN \& DNN} & \multicolumn{3}{c}{ST-ResNet \& CNN} \\
\cmidrule{2-7}
                  & CNN & DNN    & Common    & ST-ResNet & CNN   & Common\\
\midrule
Top 3 & 1752      & 251 & 116 & 2203      & 1690 & 809\\
Top 5   & 2779      & 410 & 188 & 3559      & 2698 & 1355\\
Top 10   & 5303      & 797 & 362 & 6811      & 5124 & 2750\\
\bottomrule
\end{tabular}
\label{tab1}
\end{table*}

\vspace{0.4em}
\noindent\textbf{Generality.}~We first apply CNN and ST-ResNet on \trafps, respectively. The right part of the Table~\ref{tab1} shows the statistics of the routes of top $K$ that have the most impact on the selected area found through \trafps. When selecting each region, we count all the top $K$ routes that exist in the information table. For instance, 1690 routes exist in the top 3 routes that influence the traffic when applying the CNN to predict the traffic and 2203 routes when applying the ST-ResNet. In addition, when I select the same region to interpret, 809 routes appear in top 3 simultaneously. I also find that the common routes and the proportion of the common routes in each prediction model increase as the $K$ increases. The right part of the Table~\ref{tab1} shows the statistics of applying CNN and DNN; we can find similar findings. However, the number of found routes in DNN is far less than when applying ST-ResNet and CNN. This is because the prediction principles of the two kinds of models (CNN \& ST-ResNet and DNN) are different. CNN \& ST-ResNet pay more attention to the surrounding regions, which may consider the road that is not critical (\ie, secondary). Nevertheless, the DNN encodes the whole region as a vector and may consider more on the critical roads (\ie, freeway and primary). Therefore the found routes by applying CNN and ST-ResNet include many secondary roads that surround the selected region, but in the findings in DNN, more primary roads from the global are included.

Although the findings for applying different models are not equal, the results show that our interpretation method, to some extent, is general when interpreting different traffic prediction models, which can interpret different models to similar results.

\vspace{0.4em}
\noindent\textbf{Limitation \& Future Work.}~As \textbf{E.D} and \textbf{E.E} mentioned in the expert interview, although the \trafps~can support decision-making on traffic management, they hope to provide more detailed decision-making, \eg, on the exact period to control the reversible lane. Moreover, as a post-hoc interpretation method, the recommendation from the \trafps~is mainly based on the prediction model. Therefore, one of the future directions can be to explore how to combine the model better to provide a detailed recommendation for supporting decision-making.

In addition, the division of the urban area can be an important factor that affects the result. In this research, we divide the urban area into equal size grids as the input of the prediction model and aggregate the grids based on the number of intersections into several clusters, which ignores the influence of the segmentations' shape and size on prediction results. Due to the dynamic spatial variances of the urban traffic, existing research~\cite{zeng2020revisiting} has mentioned that the shape and the size of the division region tend to affect the traffic prediction result. However, this research is more focused on the interpretation of urban traffic. Hence, we divide the urban grid with the same size and set the number of grids $38 \times 36$ as the default setting. In addition, in the regional aggregation phase, we assume that, to some extent, the number of intersections can indicate the traffic volume. Nevertheless, urban traffic is a dynamic and temporal variant. Therefore, it is not a good way just considers the intersections. Instead, we can consider more detailed information, \eg, lane capacity and occupancy. Therefore, according to this research, how to segment and aggregate the urban area by considering the urban environment to facilitate urban interpretation and decision-making may be a future direction.
% % % % % % % % % % % % % % % % % % % % % % % % % % % % % % % % % % % % % % % % % % % % % % % % % % % % % % % % % % % % % % % % % % % % % % % % % % % % % % % % % % % % % % % % % % % % % % % %
\section{Conclusion}\label{sec: conclusion}
In this paper, we have introduced a novel designed interactive visual analytics approach named \trafps~to interpret the urban traffic, including two measurements for measuring the contribution to the traffic from the surroundings and a web-based visual analytics tool for supporting multi-level analysis of the traffic interpretation. The two measurements are named \regionshap~and \trajectoryshap, which are modified from the Shapley value to measure the impact of the traffic on the surroundings. The visual design of \trafps~satisfies the requirements for the three-stage visual analytical workflow from macro to micro, which consists of three newly devised views, including a map-trajectory view, radar glyph view, and fine-grained grid view. In the case study, analysts can operate the tool on real-world data and analyze it effectively to avoid future traffic congestion and help with decision-making for urban planning. In this research, we also consider the parameter selection of the K-means algorithm, and the generality of the performance on different models and discuss the division of the grid size. The feedback from the interview with domain experts demonstrates the feasibility, usability, and effectiveness of \trafps.

% % % % % % % % % % % % % % % % % % % % % % % % % % % % % % % % % % % % % % % % % % % % % % % % % % % % % % % % % % % % % % % % % % % % % % % % % % % % % % % % % % % % % % % % % % % % % % % % % % % % % % % % % % % % 

\appendix

% \subsection*{Appendix \note{(if applicable)}}
% An appendix, if needed, is presented without numbers. If there are two or
% more appendices, they should be numbered consecutively. Equations in
% appendices should be designated differently from those in the main body of
% the manuscript, e.g., (A1), (A2), etc. In each appendix equations should be
% numbered separately.

% \subsection*{Acknowledgements}
% Acknowledgements of people,
% grants, funds, etc. should be placed in a separate section before reference
% list. The names of funding organizations should be written in full. Do not
% include acknowledgements on the title page, as a footnote to the title or
% otherwise.

\subsection*{Acknowledgements}
We thank all the domain experts interviewed in this research. We also thank the reviewers for their comments and suggestions. The work is supported in part by Grant in-Aid for Scientific Research B (22H03573) of Japan Society for the Promotion of Science (JSPS), in part by the National Natural Science Foundation of China (Grant No. 92067109, 61873119, 62211530106), in part by Shenzhen Science and Technology Program (Grant No. ZDSYS20210623092007023, GJHZ20210705141808024), and in part by the Educational Commission of Guangdong Province (Grant No. 2019KZDZX1018), in part by the National Natural Science Foundation of China (No. 62202217), in part by Guangdong Basic and Applied Basic Research Foundation (No. 2023A1515012889), and in part by Guangdong Talent Import Program (No. 2021QN02X794).
\subsection*{Declaration of competing interest}

The authors have no competing interests to declare that are relevant to the
content of this article.\\

% \begin{thebibliography}{9}

% \bibitem{1} Summa, B.; Tierny, J.; Pascucci, V. Panorama weaving: Fast and flexible
% seam processing. \textit{ACM Transactions on Graphics} Vol. 31, No. 4, Article No. 83, 2012.

% \bibitem{2} Brown, M.; Lowe, D. G. Automatic panoramic image stitching using
% invariant features. \textit{International Journal of Computer Vision} Vol. 74, No. 1, 59--73, 2007.

% \bibitem{3} Brown, M.; Lowe, D. G. Recognising panoramas. In: Proceedings of the 9th
% IEEE International Conference on Computer Vision, 1218--1225, 2003.

% \bibitem{4} Meyer, Y. \textit{Oscillating Patterns in Image Processing and Nonlinear Evolution Equations: The Fifteenth Dean Jacqueline B. Lewis Memorial Lectures}. Boston, MA, USA: American Mathematical Society, 2001.

% \bibitem{5} Prewitt, J. M. S. Object enhancement and extraction. In: \textit{Picture Processing and Psychopictorics}. Lipkin, B.;
% Rosenfeld, A. Eds. New York, NY, USA: Academic Press, 15--19, 1970.

% \bibitem{6} Lu, L.; Choi, Y.-K.; Sun, F.; Wang, W. Variational circle packing based
% on power diagram. Technical Report. The University of Hong Kong, 2011.
% Available at \url{http://vr.sdu.edu.cn/~lulin/CP\_TechReport.pdf}.

% \bibitem{7} Information on \url{http://www.adobe.com/technology/projects.html}.
% \end{thebibliography}

% for bibtex
\bibliographystyle{CVMbib}
\bibliography{refs}

\begin{thebibliography}{10}
\expandafter\ifx\csname urlstyle\endcsname\relax
  \providecommand{\doi}[1]{doi:\discretionary{}{}{}#1}\else
  \providecommand{\doi}{doi:\discretionary{}{}{}\begingroup \urlstyle{rm}\Url}\fi

\bibitem{zeng2020revisiting}
Zeng W, Lin C, Lin J, Jiang J, Xia J, Turkay C, Chen W. Revisiting the modifiable areal unit problem in deep traffic prediction with visual analytics. \emph{IEEE Transactions on Visualization and Computer Graphics}, 2020, 27(2): 839--848.

\bibitem{pi2019visual}
Pi M, Yeon H, Son H, Jang Y. Visual cause analytics for traffic congestion. \emph{IEEE transactions on visualization and computer graphics}, 2019, 27(3): 2186--2201.

\bibitem{nagy2018survey}
Nagy AM, Simon V. Survey on traffic prediction in smart cities. \emph{Pervasive and Mobile Computing}, 2018, 50: 148--163.

\bibitem{lee2019visual}
Lee C, Kim Y, Jin S, Kim D, Maciejewski R, Ebert D, Ko S. A visual analytics system for exploring, monitoring, and forecasting road traffic congestion. \emph{IEEE transactions on visualization and computer graphics}, 2019, 26(11): 3133--3146.

\bibitem{liu2016smartadp}
Liu D, Weng D, Li Y, Bao J, Zheng Y, Qu H, Wu Y. Smartadp: Visual analytics of large-scale taxi trajectories for selecting billboard locations. \emph{IEEE transactions on visualization and computer graphics}, 2016, 23(1): 1--10.

\bibitem{feng2022survey}
Feng Z, Qu H, Yang SH, Ding Y, Song J. A survey of visual analytics in urban area. \emph{Expert Systems}, 2022: e13065.

\bibitem{guo2011tripvista}
Guo H, Wang Z, Yu B, Zhao H, Yuan X. Tripvista: Triple perspective visual trajectory analytics and its application on microscopic traffic data at a road intersection. In \emph{2011 IEEE Pacific Visualization Symposium}, 2011, 163--170.

\bibitem{piringer2012alvis}
Piringer H, Buchetics M, Benedik R. Alvis: Situation awareness in the surveillance of road tunnels. In \emph{2012 IEEE Conference on Visual Analytics Science and Technology (VAST)}, 2012, 153--162.

\bibitem{chen2015survey}
Chen W, Guo F, Wang FY. A survey of traffic data visualization. \emph{IEEE Transactions on Intelligent Transportation Systems}, 2015, 16(6): 2970--2984.

\bibitem{weng2020towards}
Weng D, Zheng C, Deng Z, Ma M, Bao J, Zheng Y, Xu M, Wu Y. Towards better bus networks: A visual analytics approach. \emph{IEEE transactions on visualization and computer graphics}, 2020, 27(2): 817--827.

\bibitem{he2019interactive}
He T, Bao J, Ruan S, Li R, Li Y, He H, Zheng Y. Interactive bike lane planning using sharing bikes’ trajectories. \emph{IEEE Transactions on Knowledge and Data Engineering}, 2019, 32(8): 1529--1542.

\bibitem{short2010nonlinear}
Short MB, Bertozzi AL, Brantingham PJ. Nonlinear patterns in urban crime: Hotspots, bifurcations, and suppression. \emph{SIAM Journal on Applied Dynamical Systems}, 2010, 9(2): 462--483.

\bibitem{kraak2020cartography}
Kraak MJ, Ormeling F. \emph{Cartography: visualization of geospatial data}. 2020.

\bibitem{deng2019airvis}
Deng Z, Weng D, Chen J, Liu R, Wang Z, Bao J, Zheng Y, Wu Y. AirVis: Visual analytics of air pollution propagation. \emph{IEEE transactions on visualization and computer graphics}, 2019, 26(1): 800--810.

\bibitem{ferreira2013visual}
Ferreira N, Poco J, Vo HT, Freire J, Silva CT. Visual exploration of big spatio-temporal urban data: A study of new york city taxi trips. \emph{IEEE transactions on visualization and computer graphics}, 2013, 19(12): 2149--2158.

\bibitem{andrienko2017visual}
Andrienko G, Andrienko N, Chen W, Maciejewski R, Zhao Y. Visual analytics of mobility and transportation: State of the art and further research directions. \emph{IEEE Transactions on Intelligent Transportation Systems}, 2017, 18(8): 2232--2249.

\bibitem{zheng2016visual}
Zheng Y, Wu W, Chen Y, Qu H, Ni LM. Visual analytics in urban computing: An overview. \emph{IEEE Transactions on Big Data}, 2016, 2(3): 276--296.

\bibitem{kamw2019urban}
Kamw F, Al-Dohuki S, Zhao Y, Eynon T, Sheets D, Yang J, Ye X, Chen W. Urban structure accessibility modeling and visualization for joint spatiotemporal constraints. \emph{IEEE Transactions on Intelligent Transportation Systems}, 2019, 21(1): 104--116.

\bibitem{feng2020topology}
Feng Z, Li H, Zeng W, Yang SH, Qu H. Topology density map for urban data visualization and analysis. \emph{IEEE transactions on visualization and computer graphics}, 2020, 27(2): 828--838.

\bibitem{zeng2019route}
Zeng W, Shen Q, Jiang Y, Telea A. Route-Aware Edge Bundling for Visualizing Origin-Destination Trails in Urban Traffic. \emph{Computer Graphics Forum}, 2019, 38(3): 581--593.

\bibitem{zeng2014visualizing}
Zeng W, Fu CW, Arisona SM, Erath A, Qu H. Visualizing mobility of public transportation system. \emph{IEEE transactions on visualization and computer graphics}, 2014, 20(12): 1833--1842.

\bibitem{kruger2018visual}
Kr{\"u}ger R, Simeonov G, Beck F, Ertl T. Visual interactive map matching. \emph{IEEE transactions on visualization and computer graphics}, 2018, 24(6): 1881--1892.

\bibitem{al2016semantictraj}
Al-Dohuki S, Wu Y, Kamw F, Yang J, Li X, Zhao Y, Ye X, Chen W, Ma C, Wang F. Semantictraj: A new approach to interacting with massive taxi trajectories. \emph{IEEE transactions on visualization and computer graphics}, 2016, 23(1): 11--20.

\bibitem{huang2019natural}
Huang Z, Zhao Y, Chen W, Gao S, Yu K, Xu W, Tang M, Zhu M, Xu M. A natural-language-based visual query approach of uncertain human trajectories. \emph{IEEE Transactions on Visualization and Computer Graphics}, 2019, 26(1): 1256--1266.

\bibitem{wu2015telcovis}
Wu W, Xu J, Zeng H, Zheng Y, Qu H, Ni B, Yuan M, Ni LM. Telcovis: Visual exploration of co-occurrence in urban human mobility based on telco data. \emph{IEEE transactions on visualization and computer graphics}, 2015, 22(1): 935--944.

\bibitem{shen2017streetvizor}
Shen Q, Zeng W, Ye Y, Arisona SM, Schubiger S, Burkhard R, Qu H. StreetVizor: Visual exploration of human-scale urban forms based on street views. \emph{IEEE Transactions on Visualization and Computer Graphics}, 2017, 24(1): 1004--1013.

\bibitem{qu2007visual}
Qu H, Chan WY, Xu A, Chung KL, Lau KH, Guo P. Visual analysis of the air pollution problem in Hong Kong. \emph{IEEE Transactions on visualization and Computer Graphics}, 2007, 13(6): 1408--1415.

\bibitem{cao2017voila}
Cao N, Lin C, Zhu Q, Lin YR, Teng X, Wen X. Voila: Visual anomaly detection and monitoring with streaming spatiotemporal data. \emph{IEEE transactions on visualization and computer graphics}, 2017, 24(1): 23--33.

\bibitem{weng2018homefinder}
Weng D, Zhu H, Bao J, Zheng Y, Wu Y. Homefinder revisited: Finding ideal homes with reachability-centric multi-criteria decision making. In \emph{Proceedings of the 2018 CHI Conference on Human Factors in Computing Systems}, 2018, 1--12.

\bibitem{li2020warehouse}
Li Q, Liu Q, Tang C, Li Z, Wei S, Peng X, Zheng M, Chen T, Yang Q. Warehouse Vis: A visual analytics approach to facilitating warehouse location selection for business districts. \emph{Computer Graphics Forum}, 2020, 39(3): 483--495.

\bibitem{di2015allaboard}
Di~Lorenzo G, Sbodio M, Calabrese F, Berlingerio M, Pinelli F, Nair R. Allaboard: visual exploration of cellphone mobility data to optimise public transport. \emph{IEEE transactions on visualization and computer graphics}, 2015, 22(2): 1036--1050.

\bibitem{weng2020pareto}
Weng D, Chen R, Zhang J, Bao J, Zheng Y, Wu Y. Pareto-optimal transit route planning with multi-objective monte-carlo tree search. \emph{IEEE Transactions on Intelligent Transportation Systems}, 2020, 22(2): 1185--1195.

\bibitem{ming2017understanding}
Ming Y, Cao S, Zhang R, Li Z, Chen Y, Song Y, Qu H. Understanding hidden memories of recurrent neural networks. In \emph{2017 IEEE Conference on Visual Analytics Science and Technology (VAST)}, 2017, 13--24.

\bibitem{ma2019explaining}
Ma Y, Xie T, Li J, Maciejewski R. Explaining vulnerabilities to adversarial machine learning through visual analytics. \emph{IEEE transactions on visualization and computer graphics}, 2019, 26(1): 1075--1085.

\bibitem{ma2020visual}
Ma Y, Fan A, He J, Nelakurthi AR, Maciejewski R. A visual analytics framework for explaining and diagnosing transfer learning processes. \emph{IEEE Transactions on Visualization and Computer Graphics}, 2020, 27(2): 1385--1395.

\bibitem{chen2017vaud}
Chen W, Huang Z, Wu F, Zhu M, Guan H, Maciejewski R. VAUD: A visual analysis approach for exploring spatio-temporal urban data. \emph{IEEE transactions on visualization and computer graphics}, 2017, 24(9): 2636--2648.

\bibitem{abdul2018trends}
Abdul A, Vermeulen J, Wang D, Lim BY, Kankanhalli M. Trends and trajectories for explainable, accountable and intelligible systems: An hci research agenda. In \emph{Proceedings of the 2018 CHI conference on human factors in computing systems}, 2018, 1--18.

\bibitem{liu2017towards}
Liu S, Wang X, Liu M, Zhu J. Towards better analysis of machine learning models: A visual analytics perspective. \emph{Visual Informatics}, 2017, 1(1): 48--56.

\bibitem{guo2020dynamic}
Guo K, Hu Y, Qian Z, Sun Y, Gao J, Yin B. Dynamic graph convolution network for traffic forecasting based on latent network of laplace matrix estimation. \emph{IEEE Transactions on Intelligent Transportation Systems}, 2020.

\bibitem{zhang2017deep}
Zhang J, Zheng Y, Qi D. Deep spatio-temporal residual networks for citywide crowd flows prediction. In \emph{Proceedings of the AAAI conference on artificial intelligence}, volume 31(1), 2017, 1.

\bibitem{wang2013visual}
Wang Z, Lu M, Yuan X, Zhang J, Van De~Wetering H. Visual traffic jam analysis based on trajectory data. \emph{IEEE transactions on visualization and computer graphics}, 2013, 19(12): 2159--2168.

\bibitem{shin2021rcmvis}
Shin D, Jo J, Kim B, Song H, Cho SH, Seo J. RCMVis: A Visual Analytics System for Route Choice Modeling. \emph{IEEE Transactions on Visualization and Computer Graphics}, 2021.

\bibitem{gou2020vatld}
Gou L, Zou L, Li N, Hofmann M, Shekar AK, Wendt A, Ren L. VATLD: a visual analytics system to assess, understand and improve traffic light detection. \emph{IEEE transactions on visualization and computer graphics}, 2020, 27(2): 261--271.

\bibitem{he2021can}
He W, Zou L, Shekar AK, Gou L, Ren L. Where can we help? a visual analytics approach to diagnosing and improving semantic segmentation of movable objects. \emph{IEEE Transactions on Visualization and Computer Graphics}, 2021, 28(1): 1040--1050.

\bibitem{doshi2017towards}
Doshi-Velez F, Kim B. Towards a rigorous science of interpretable machine learning. \emph{arXiv preprint arXiv:1702.08608}, 2017.

\bibitem{arrieta2020explainable}
Arrieta AB, D{\'\i}az-Rodr{\'\i}guez N, Del~Ser J, Bennetot A, Tabik S, Barbado A, Garc{\'\i}a S, Gil-L{\'o}pez S, Molina D, Benjamins R, et~al.. Explainable Artificial Intelligence (XAI): Concepts, taxonomies, opportunities and challenges toward responsible AI. \emph{Information fusion}, 2020, 58: 82--115.

\bibitem{ribeiro2018anchors}
Ribeiro MT, Singh S, Guestrin C. Anchors: High-precision model-agnostic explanations. In \emph{Proceedings of the AAAI conference on artificial intelligence}, volume~32, 2018, 1.

\bibitem{ribeiro2016should}
Ribeiro MT, Singh S, Guestrin C. ``Why should i trust you?'' Explaining the predictions of any classifier. In \emph{Proceedings of the 22nd ACM SIGKDD International Conference on Knowledge Discovery and Data Mining}, 2016, 1135--1144.

\bibitem{lundberg2017unified}
Lundberg SM, Lee SI. A unified approach to interpreting model predictions. \emph{Advances in Neural Information Processing Systems}, 2017, 30.

\bibitem{zhang2021building}
Zhang D, Zhou H, Zhang H, Bao X, Huo D, Chen R, Cheng X, Wu M, Zhang Q. Building interpretable interaction trees for deep nlp models. In \emph{Proceedings of the AAAI Conference on Artificial Intelligence}, volume 35(16), 2021, 14328--14337.

\bibitem{sundararajan2017axiomatic}
Sundararajan M, Taly A, Yan Q. Axiomatic attribution for deep networks. In \emph{International Conference on Machine Learning}, 2017, 3319--3328.

\bibitem{parsa2020toward}
Parsa AB, Movahedi A, Taghipour H, Derrible S, Mohammadian AK. Toward safer highways, application of XGBoost and SHAP for real-time accident detection and feature analysis. \emph{Accident Analysis \& Prevention}, 2020, 136: 105405.

\bibitem{parsa2021data}
Parsa AB, Shabanpour R, Mohammadian A, Auld J, Stephens T. A data-driven approach to characterize the impact of connected and autonomous vehicles on traffic flow. \emph{Transportation Letters}, 2021, 13(10): 687--695.

\bibitem{rizzo2019reinforcement}
Rizzo SG, Vantini G, Chawla S. Reinforcement learning with explainability for traffic signal control. In \emph{2019 IEEE Intelligent Transportation Systems Conference (ITSC)}, 2019, 3567--3572.

\bibitem{ishikawa1990introduction}
Ishikawa K, Loftus JH. \emph{Introduction to quality control}. 1990.

\bibitem{zhang2016dnn}
Zhang J, Zheng Y, Qi D, Li R, Yi X. DNN-based prediction model for spatio-temporal data. In \emph{Proceedings of the 24th ACM SIGSPATIAL international conference on advances in geographic information systems}, 2016, 1--4.

\bibitem{zhang2019flow}
Zhang J, Zheng Y, Sun J, Qi D. Flow prediction in spatio-temporal networks based on multitask deep learning. \emph{IEEE Transactions on Knowledge and Data Engineering}, 2019, 32(3): 468--478.

\bibitem{munzner2009nested}
Munzner T. A nested model for visualization design and validation. \emph{IEEE transactions on visualization and computer graphics}, 2009, 15(6): 921--928.

\bibitem{shi2020urbanmotion}
Shi L, Huang C, Liu M, Yan J, Jiang T, Tan Z, Hu Y, Chen W, Zhang X. Urbanmotion: Visual analysis of metropolitan-scale sparse trajectories. \emph{IEEE Transactions on Visualization and Computer Graphics}, 2020, 27(10): 3881--3899.

\end{thebibliography}

\newpage
\subsection*{Author biography}
% \note{(at least the first author's and the corresponding author's)}

\begin{biography}[{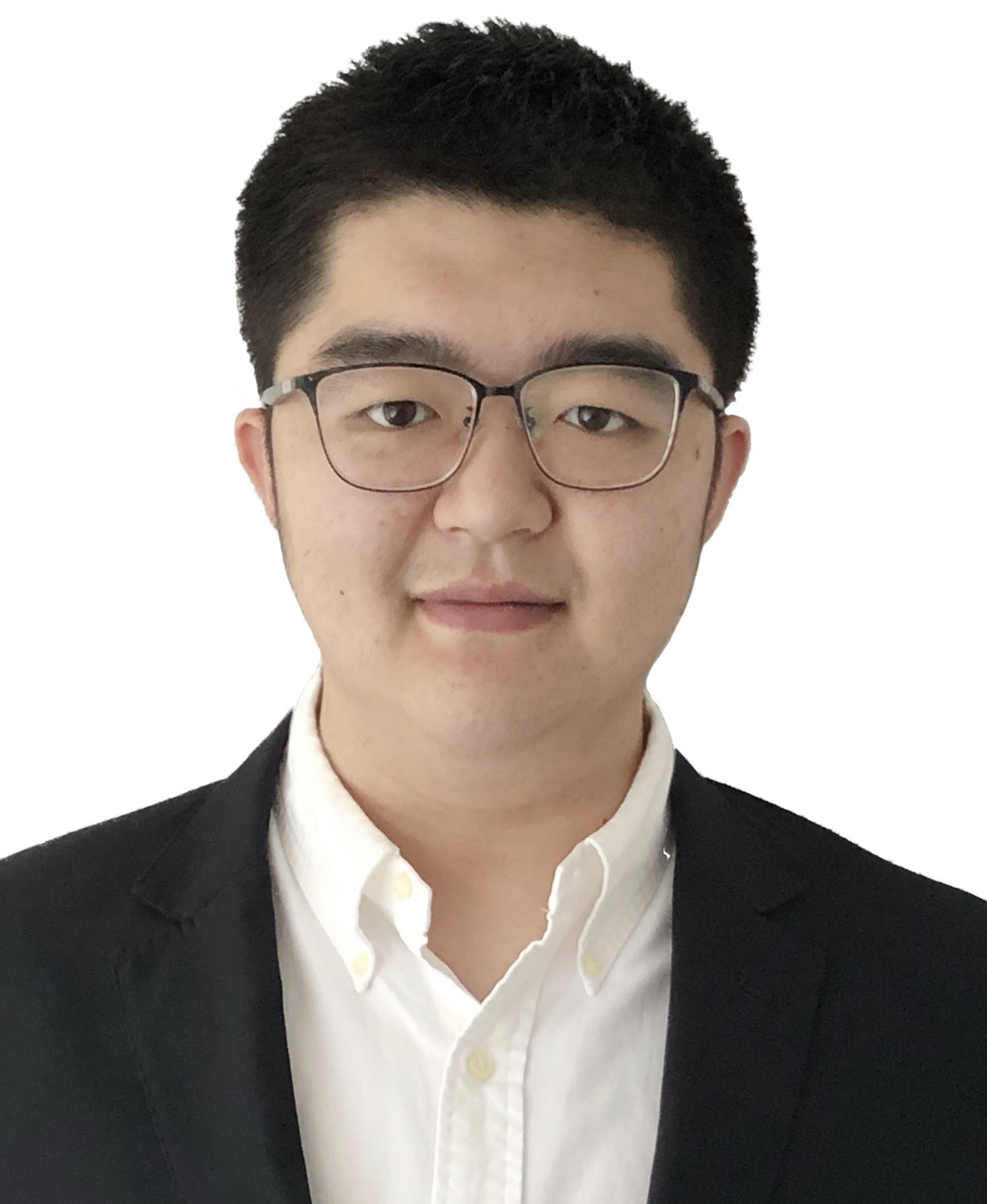}]{Zezheng Feng} received a B.E. degree from Northeastern University (NEU), China, in 2017, and an M.S. degree (with distinction) from Loughborough University, UK, in 2018. He is currently a Ph.D. candidate in the Department of Computer Science and Engineering (CSE) at the Hong Kong University of Science and Technology (HKUST) and sponsored by a joint Ph.D. program between HKUST and Southern University of Science and Technology (SUSTech). His recent research interests include visualization and visual analytics, explainable artificial intelligence (XAI), and urban computing. For more information, please visit \url{https://jerrodfeng.github.io/}
\end{biography}

\vspace*{2em}

\begin{biography}[{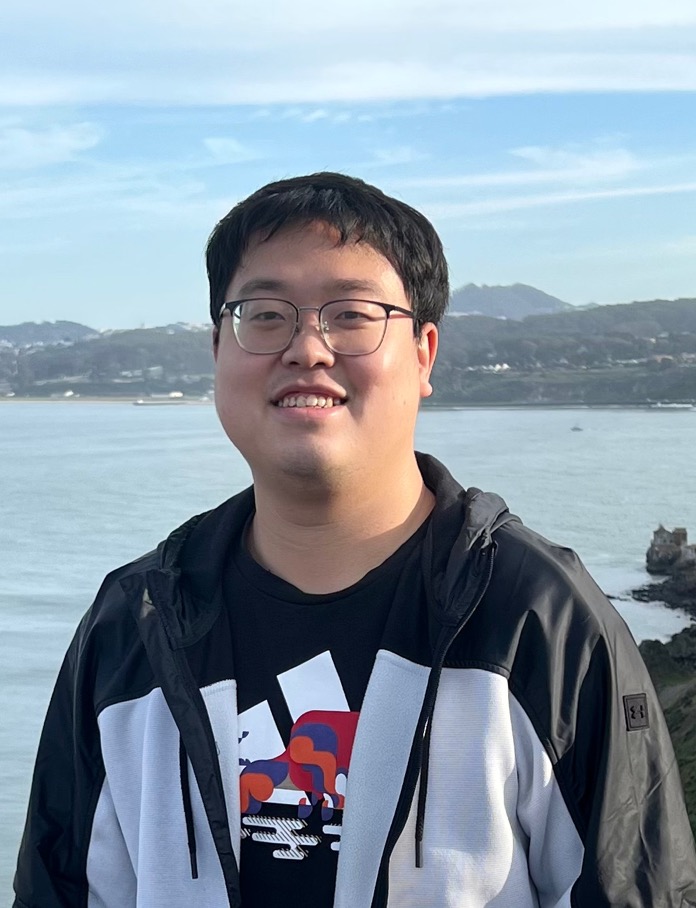}]{Yifan Jiang} is working toward a Ph.D. degree in computer science from USC. He received a B.E. degree from the Southern University of Science and Technology in 2021. His research interests are machine learning explanation, natural language processes, commonsense reasoning, and data visualization.
\end{biography}

\vspace*{1em}

\begin{biography}[{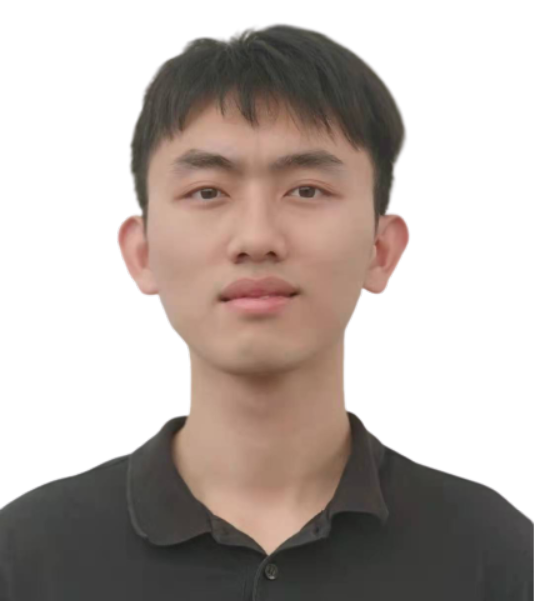}]{Hongjun Wang} 
is working toward an M.S. degree in  computer science and technology from the Southern University of Science and Technology, China. He received a B.E. degree from the Nanjing University of Posts and Telecommunications, China, in 2019. His research interests are broadly in machine learning, urban computing, explainable AI, data mining, and data visualization.
\end{biography}

\vspace*{1em}

\begin{biography}[{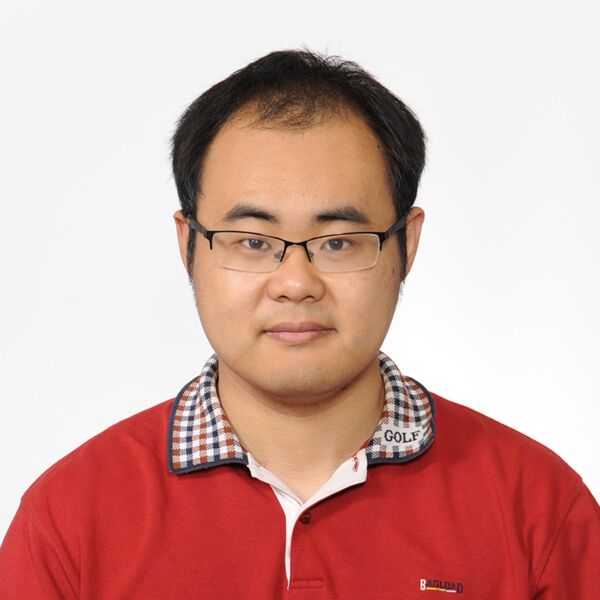}]{Zipei Fan} 
is a project lecturer at the University of Tokyo. He graduated from School of Computer Science and Engineering of Beihang University in 2012 and received the master and doctoral degree on Civil Engineering from the University of Tokyo in 2014 and 2017 respectively. His research interests include data mining, Internet of Things, machine learning and applications on smart city. He has published more than 40 papers in journals and conferences including TKDE, IMWUT/UbiComp, WWWJ, IJCAI, CIKM, SIGSPATIAL and etc, and have been invited as reviewers for conferences and journals such as: IJCAI, AAAI, ECML, TKDE, UbiComp, Transactions on Mobile Computing (TMC), WWWJ, Transactions on Big Data (TBD), etc.

\end{biography}

\vspace*{1em}

\begin{biography}[{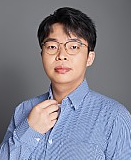}]{Yuxin Ma} is a tenure-track assistant professor in the Department of Computer Science and Engineering, Southern University of Science and Technology (SUSTech), China. He received his B.Eng. and Ph.D. degrees from Zhejiang University. Before joining SUSTech, he worked as a Postdoctoral Research Associate in VADER Lab, CIDSE, Arizona State University. His primary research interests are in the areas of visualization and visual analytics, focusing on explainable AI, high-dimensional data, and spatiotemporal data.

\end{biography}

\vspace*{1em}

\begin{biography}[{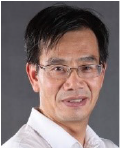}]{Shuang-Hua Yang} received the B.S. degree in instrument and automation and the M.S. degree in process control from the China University of Petroleum (Huadong), Beijing, China, in 1983 and 1986, respectively, and the Ph.D. degree in intelligent systems from Zhejiang University, Hangzhou, China, in 1991. He is currently the director of the Shenzhen Key Laboratory of Safety and Security for Next Generation of Industrial Internet at the Southern University of Science and Technology, China, and also the Head of Department of Computer Science at the University of Reading, UK. His research interests include cyber-physical systems, the Internet of Things, wireless network-based monitoring and control, and safety-critical systems. He is a fellow of IET and InstMC, U.K. He is also an Associate Editor of IET Cyber-Physical Systems: Theory and Applications.

\end{biography}

\vspace*{1em}

\begin{biography}[{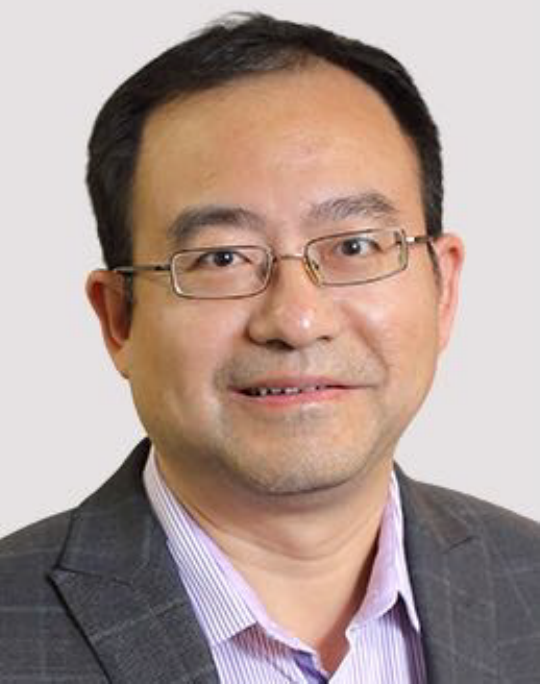}]{Huamin Qu} is a professor in the Department of Computer Science and Engineering (CSE) at the Hong Kong University of Science and Technology (HKUST) and also the director of the interdisciplinary program office (IPO) of HKUST. He obtained a BS in Mathematics from Xi'an Jiaotong University, China, an MS and a PhD in Computer Science from the Stony Brook University. His main research interests are in visualization and human-computer interaction, with focuses on urban informatics, social network analysis, E-learning, text visualization, and explainable artificial intelligence (XAI). For more information, please visit \url{http://huamin.org/}.

\end{biography}

\vspace*{1em}

\begin{biography}[{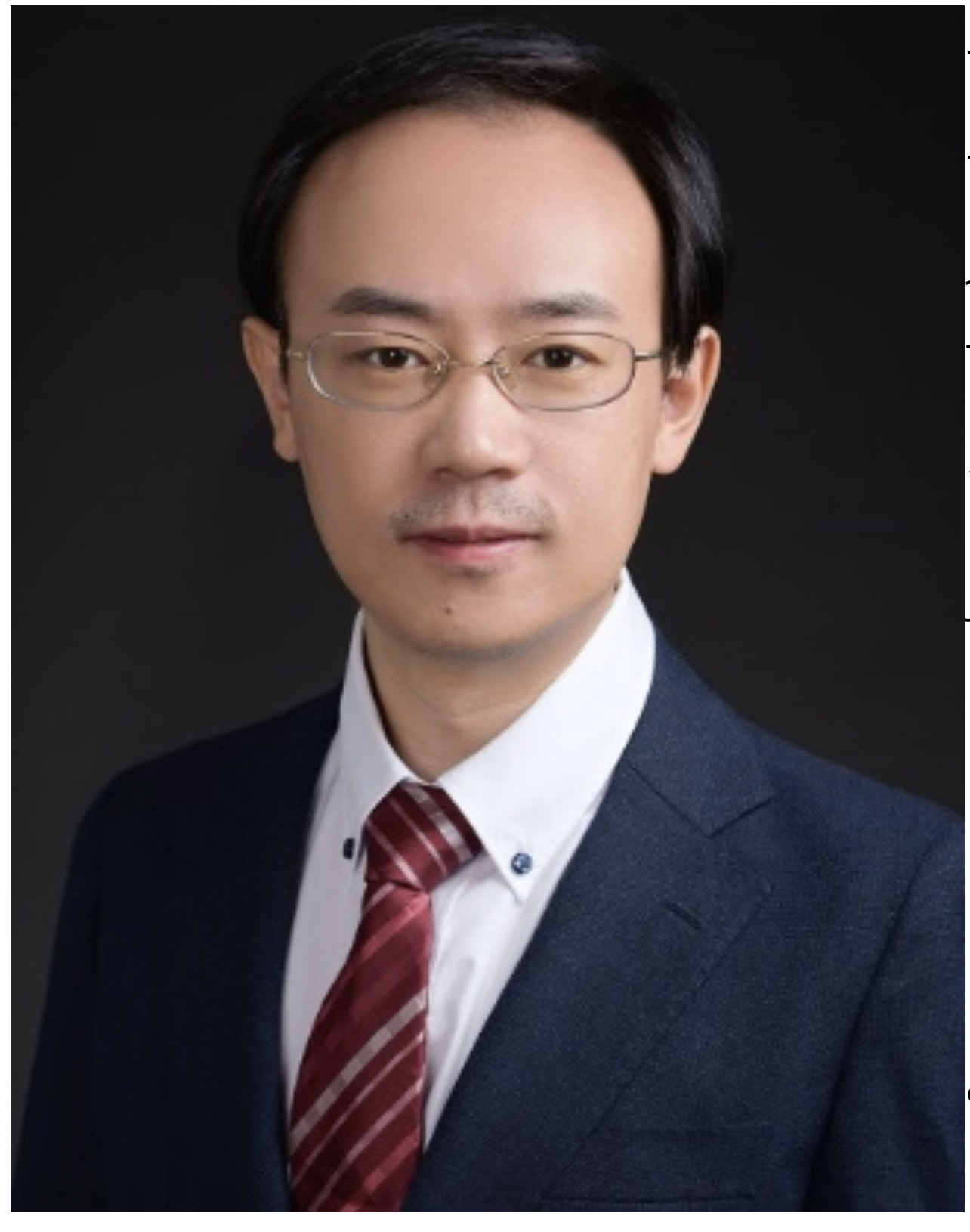}]{Xuan Song} received the Ph.D. degree in signal and information processing from Peking University in 2010. In 2017, he was selected as an Excellent Young Researcher of Japan MEXT. In the past ten years, he led and participated in many important projects as a principal investigator or primary actor in Japan, such as the DIAS/GRENE Grant of MEXT, Japan; Japan/US Big Data and Disaster Project of JST, Japan; Young Scientists Grant and Scientific Research Grant of MEXT, Japan; Research Grant of MLIT, Japan; CORE Project of Microsoft; Grant of JR EAST Company and Hitachi Company, Japan. He served as Associate Editor, Guest Editor, Area Chair, Program Committee Member or reviewer for many famous journals and top-tier conferences, such as IMWUT, IEEE Transactions on Multimedia, WWW Journal, Big Data Journal, ISTC, MIPR, ACM TIST, IEEE TKDE, UbiComp, ICCV, CVPR, ICRA and etc.

\end{biography}

\vspace*{2.6em}

% \subsection*{Graphical abstract}

% Graphical abstract is optional yet highly encouraged to supply which can
% summarize your content vividly in one picture (at least 600 dpi, 5 cm
% $\times $ 8 cm, the ratio of height to length should be less than 1 and
% larger than 5/8). We will upload it onto SpringerLink and display it on the
% webpage of this paper.

\end{document}